\documentclass[nohyperref]{article}

\usepackage{microtype}
\usepackage{graphicx}
\usepackage{booktabs} % for professional tables

\usepackage{hyperref}

\usepackage[accepted]{icml2023}

\usepackage{amsmath}
\usepackage{amssymb}
\usepackage{mathtools}
\usepackage{amsthm}

\usepackage[capitalize,noabbrev]{cleveref}

\usepackage{longtable}

\usepackage{subcaption}
\usepackage[export]{adjustbox}
\usepackage{caption}
\usepackage{tikz}
\usetikzlibrary{bayesnet}
\usepackage{multirow}
\usepackage{longtable}
\usepackage{makecell}
\usepackage{tikzscale}

\theoremstyle{plain}

\theoremstyle{definition}

\theoremstyle{remark}

\usepackage[textsize=tiny]{todonotes}

\newcommand{\cblock}[3]{
 \begin{tikzpicture}
   [
   node/.style={square, minimum size=10mm, thick, line width=0pt, anchor=center, align=center},
   ]
   \node[fill={rgb,255:red,#1;green,#2;blue,#3}] () [] {};
 \end{tikzpicture}%
}

\newcommand{\cstar}[3]{
  \begin{tikzpicture}[scale=0.5]
    \node[star,star points=5,star point ratio=2.5,          minimum size=10mm,thick,          fill={rgb,255:red,#1;green,#2;blue,#3}] () [] {};
  \end{tikzpicture}%
}

\usepackage[textsize=tiny]{todonotes}

\icmltitlerunning{Training Language Models with Language Feedback at Scale}

\begin{document}
\twocolumn[
\icmltitle{Training Language Models with Language Feedback at Scale}

% It is OKAY to include author information, even for blind
% submissions: the style file will automatically remove it for you
% unless you've provided the [accepted] option to the icml2023
% package.

% List of affiliations: The first argument should be a (short)
% identifier you will use later to specify author affiliations
% Academic affiliations should list Department, University, City, Region, Country
% Industry affiliations should list Company, City, Region, Country

% You can specify symbols, otherwise they are numbered in order.
% Ideally, you should not use this facility. Affiliations will be numbered
% in order of appearance and this is the preferred way.
\icmlsetsymbol{equal}{*}

\begin{icmlauthorlist}
\icmlauthor{Jérémy Scheurer}{nyu,far}
\icmlauthor{Jon Ander Campos}{nyu,ba}
\icmlauthor{Tomasz Korbak}{nyu,far,su}
\icmlauthor{Jun Shern Chan}{nyu,far}
\icmlauthor{Angelica Chen}{nyu}
\icmlauthor{Kyunghyun Cho}{nyu,gen,cif}
\icmlauthor{Ethan Perez}{nyu,far,ant}
\end{icmlauthorlist}

\icmlaffiliation{nyu}{New York University}
\icmlaffiliation{far}{FAR AI}
\icmlaffiliation{ba}{HiTZ Center, University of the Basque Country UPV/EHU}
\icmlaffiliation{su}{University of Sussex}
\icmlaffiliation{gen}{Genentech}
\icmlaffiliation{cif}{CIFAR LMB}
\icmlaffiliation{ant}{Anthropic}

\icmlcorrespondingauthor{Jérémy Scheurer}{jeremy.scheurer@nyu.edu}
\icmlcorrespondingauthor{Ethan Perez}{ethan@anthropic.com}

% You may provide any keywords that you
% find helpful for describing your paper; these are used to populate
% the "keywords" metadata in the PDF but will not be shown in the document
\icmlkeywords{Language Models, Bayesian Inference, Reinforcement Learning from Human Feedback}

\vskip 0.3in
]

% this must go after the closing bracket ] following \twocolumn[ ...

% This command actually creates the footnote in the first column
% listing the affiliations and the copyright notice.
% The command takes one argument, which is text to display at the start of the footnote.
% The \icmlEqualContribution command is standard text for equal contribution.
% Remove it (just {}) if you do not need this facility.

%\printAffiliationsAndNotice{}  % leave blank if no need to mention equal contribution
%\printAffiliationsAndNotice{\icmlEqualContribution} % otherwise use the standard text.

\printAffiliationsAndNotice{}

\begin{abstract}
Pretrained language models often generate outputs that are not in line with human preferences, such as harmful text or factually incorrect summaries. Recent work approaches the above issues by learning from a simple form of human feedback: comparisons between pairs of model-generated outputs. However, comparison feedback only conveys limited information about human preferences. In this paper, we introduce Imitation learning from Language Feedback (ILF), a new approach that utilizes more informative language feedback. ILF consists of three steps that are applied iteratively: first, conditioning the language model on the input, an initial LM output, and feedback to generate refinements. Second, selecting the refinement incorporating the most feedback. Third, finetuning the language model to maximize the likelihood of the chosen refinement given the input. We show theoretically that ILF can be viewed as Bayesian Inference, similar to Reinforcement Learning from human feedback. We evaluate ILF's effectiveness on a carefully-controlled toy task and a realistic summarization task.
Our experiments demonstrate that large language models accurately incorporate feedback and that finetuning with ILF scales well with the dataset size, even outperforming finetuning on human summaries. Learning from both language and comparison feedback outperforms learning from each alone, achieving human-level summarization performance. 
\end{abstract}

\begin{figure}[ht]
    \centering \includegraphics[width=\columnwidth]{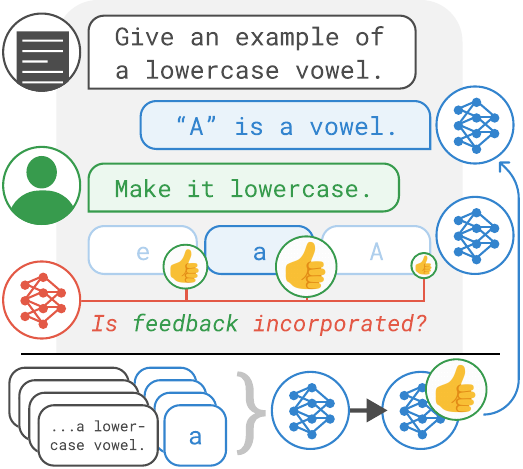}
    \caption{To learn from language feedback on a language model (LM) output, we have an LM generate multiple refinements of the original output based on the feedback. We use an LM to pick the best refinement and finetune the original LM to maximize the likelihood of the chosen refinement.}
    \label{fig:illustration}
\end{figure}

\section{Introduction}
Language Models (LMs) achieve strong performance across diverse NLP tasks, from summarization to question answering and dialog \citep[][\textit{inter alia}]{Radford2018ImprovingLU,Radford2019LanguageMA,brown2020language, rae2021scaling}. One of their key limitations, however, is that they generate text that violates human preferences, such as misinformation \cite{lin2021truthfulqa}, offensive language \cite{gehman2020realtoxicityprompts}, and factually incorrect summaries \cite{stiennon2020learning}. To alleviate such issues, existing methods train LMs to generate text that scores highly according to human preferences or a predictive model thereof~\cite{ziegler2019fine,stiennon2020learning, nakano2021webgpt,ouyang2022training}. These approaches learn from human feedback regarding which of two outputs is better. However, each comparison only conveys limited information about human preferences.

We propose an alternative approach that learns from language feedback, an information-rich and natural form of human feedback.
We introduce Imitation learning from Language Feedback (ILF), a 3-step algorithm for learning from language feedback (Fig.~\ref{fig:illustration}). First, we generate multiple refinements of an LM-generated output, given the input, initial LM-generated output, and human-written feedback on the output. Second, we use an instruction-finetuned LM to choose the refinement that best incorporates the feedback. Third, we finetune the LM that generated the initial output on the chosen refinement given the input. In this way, we finetune an LM using language feedback; with the resulting model, we may then collect more feedback on its outputs and learn with the above refine-and-finetune approach. The algorithm's pseudocode (Algorithm~\ref{alg:expert_iteration}) and the corresponding graphical model are shown in Fig~\ref{fig:data_generating_process}.
ILF departs from prior work, which uses reinforcement learning (RL)~\citep[][\textit{inter alia}]{ziegler2019fine, stiennon2020learning} or auxiliary losses~\cite{stacey2021natural} and cannot be straightforwardly generalized to using free-form language feedback.

\begin{figure}
\centering
\begin{minipage}{0.48\textwidth}
   % \vspace{-12px}
    \begin{subfigure}{0.2\textwidth}
           \scalebox{0.9}{\begin{tikzpicture}
  % Define nodes
  \node[obs]           (c) {$c$};
  \node[latent, right=0.6cm of c]       (x_1) {$x_1$};
  %\node[latent, right=0.6cm of x_1]        (I) {$\mathcal{I}$};

  % Connect the nodes
  \edge {c} {x_1} ; %
  %\edge {x_1} {I} ; %
  %\draw [->] (c) to [out=40,in=140] (I);

\end{tikzpicture}}
         %\label{fig:dgp_p}
    \end{subfigure}
\hspace{0.12\textwidth}%
    \begin{subfigure}{0.2\textwidth}
        \scalebox{0.9}{\begin{tikzpicture}
  % Define nodes
  \node[obs]           (c) {$c$};
  \node[latent, right=0.6cm of c]       (x_0) {$x_0$};
  \node[latent, right=0.6cm of x_0]     (f) {$f$};
  \node[latent, right=0.6cm of f]       (x_1) {$x_1$};
  %\node[obs, right=0.6cm of x_1]        (I) {$\mathcal{I}$};

  % Connect the nodes
  \edge {c} {x_0} ; %
  \edge {x_0} {f} ; %
  \edge {f} {x_1} ; %
  %\edge {x_1} {I} ; %
  \draw [->] (c) to [out=40,in=140] (f);
  \draw [->] (c) to [out=40,in=140] (x_1);
  %\draw [->] (c) to [out=40,in=140] (I);
  \draw [->] (x_0) to [out=40,in=140] (x_1);
  %\draw [->] (x_0) to [out=40,in=140] (I);
  \draw [->] (f) to [out=40,in=140] (x_1);
  %\draw [->] (f) to [out=40,in=140] (I);
\end{tikzpicture}}
        % \label{fig:dgp_q}  
\end{subfigure}
\end{minipage}
\setcounter{figure}{1}    
\begin{minipage}{0.48\textwidth}
\begin{algorithm}[H]
 \caption{Imitation Learning from Language Feedback}
   \label{alg:expert_iteration}
\begin{algorithmic}
\STATE {\bfseries Input:} number of iterations $K$, a sequence of sets of source documents $\mathcal{C}=[\mathcal{C}_1, ..., \mathcal{C}_K]$, language model $\pi_{\theta}$, refinement language model $\pi_{\psi}$, reward model $R$
    \FOR{$k$ {\bfseries in} $1...K$}
        \STATE Initialize finetuning dataset $\mathcal{D}_k = \{ \}$
       \FOR{document $c$ {\bfseries in} $\mathcal{C}_k$}
           \STATE $x_0 \sim \pi_{\theta}(x_0|c)$
           \STATE Human provides feedback $f$ on $(c,x_0)$
           \STATE $\{x_1^1, \dots, x_1^N \} \sim \pi_{\psi}(x_1|c,x_0,f)$
           \STATE $x_1 = \text{argmax}_{x_1^n} 
          R(x_1^i|x_0, f, c)$
            \STATE Add $(c,x_1)$ to $\mathcal{D}_k$
        \ENDFOR
        \STATE Update $\pi_{\theta}$ by supervised finetuning on $\mathcal{D}_k$ (as in Eq.~\ref{eq:final_objective})
   \ENDFOR
\end{algorithmic}
\end{algorithm}
\end{minipage}
\caption{
 \textbf{Top Left:} The graphical model of the target distribution $p_{\theta}$ that our algorithm approximates. $c$ is a context and $x_1$ is a high-quality LM output. \textbf{Top Right:} Graphical model of the proposal distribution $q$ for importance sampling. $x_0$ is an initial LM output and $f$ is language feedback on $x_0$. \textbf{Bottom:} Pseudocode for our learning algorithm.} 
 \label{fig:data_generating_process}
 \vspace{-10px}
 \end{figure}

We analyze our approach both theoretically and empirically.
We show that ILF can be viewed as Bayesian Inference, similar to RL with Human Feedback with KL penalties~\citep{korbak2022rl}.
We then validate our algorithm on a carefully-controlled synthetic task of removing offensive words from a sentence with GPT-3-based models~\citep{brown2020language,ouyang2022training}.
We find that only the largest GPT-3-based models (175B parameters) accurately refine outputs.
Using this insight, we use the largest GPT-3 models to test our algorithm on text summarization, following~\citet{stiennon2020learning}.
% A model trained with our algorithm generates summaries that human evaluators prefer to human reference summaries $\sim$51\% of the time.
Our work extends our earlier unpublished results~\citep{scheurer2022training}, showing that ILF improves LM-generated summaries monotonically with the amount of feedback provided, testing up to 5k samples.
In all data regimes, ILF leads to comparable or better results to finetuning on \textit{human-written} summaries, suggesting our approach is a strong alternative to supervised learning on human demonstrations.
We also introduce an approach for learning from both language and comparison feedback by choosing the best-of-N samples from an ILF-trained model using a model trained with comparison feedback.
The hybrid approach outperforms learning from each form of feedback alone, leading to summaries that human evaluators prefer over high-quality human reference summaries $\sim 50.8$\% of the time.
% We find that combining ILF with a reward model trained on binary comparisons produces the best results and achieves roughly human-level summarization performance.
Our analysis shows that LM-generated refinements typically incorporate the feedback, especially when we use an LM to choose the refinement that best incorporates the feedback.
In our concurrent paper~\citep{chen2023feedback}, we show that ILF also achieves strong performance on code generation.
Our results suggest that language feedback is a promising avenue for learning human preferences.

\section{Methods}
\label{sec:methods}
We now formulate the problem setting and describe our approach. We aim to generate improved outputs $x_1$ (e.g., high-quality summaries), according to human preferences, given language feedback $f$ on an initial model-generated output $x_0$, and a context $c$ (e.g., a source document). We tackle this problem by updating an LM $\pi_{\theta}$ based on evidence provided by language feedback. 

Our goal is to sample a diverse set of high-quality outputs $x_1$ given a context $c$ (e.g., a summary of a document), where $c$ is drawn from the context distribution $p(c)$. We do so by fitting an autoregressive LM $\pi_{\theta}$ to approximate the ground-truth distribution $p_c^*(x_1)$ which is proportional to the quality of $x_1$, measured by the reward function $R$. Fitting $\pi_{\theta}$ can be written down as minimizing the expected KL-divergence from the true distribution $p_c^*(x_1)$ to $\pi_{\theta}$ over the context distribution $p(c)$:
% We use an LM $\pi_{\theta}$ to generate an output $x_1$, by conditioning on the context $c$, i.e., $x_1 \sim p_{\theta}(x_1|c)$. We can thus formalize our objective as minimizing the KL-divergence between the target distribution $p_c^*(x1)$ under the expectation of a context distribution $p(c)$: 

\begin{align}
\label{eq:objective}
\min_{\theta} \mathbb{E}_{c\sim p(c)}&\mathrm{KL}(p^*_c, \pi_\theta), \\
\text{where } p_c^*(x_1) &\propto \exp(\beta R(x_1|c)). \nonumber 
\end{align}

Minimizing the objective in Eq.~\ref{eq:objective} equivalent to minimizing the cross-entropy loss (i.e., supervised learning):
\begin{align*}
\mathcal{L}(\theta) &=-\mathbb{E}_{c\sim p(c)}\mathcal{L}_\theta(c), \\
\text{where } \mathcal{L}_\theta(c) &= \sum_{x_1} p^*_c(x_1) \log \pi_\theta (x_1|c). \nonumber
\end{align*}

It is intractable to compute this loss exactly for a number of reasons, including the exponential size of the space of $x_1$ as well as the intractability of computing the normalization constant of $p_c^*(x_1)$. To avoid the first issue, we use Monte Carlo approximation sampling using a small set of samples drawn from $p_c^*$. Directly sampling from $p_c^*$ is however still intractable. We thus resort to using importance sampling with a proposal distribution $q_c(x_1)$ that is simpler to sample:

\begin{align}
\mathcal{L}_\theta(c) =\sum_{x_1} q_c(x_1) \frac{p^*_c(x_1)}{q_c(x_1)} \log \pi_\theta(x_1|c)  \label{eq:importance_sampling}
\end{align}

 To minimize the variance, we must design $q_c$ to be as close as possible to $p_c^*$. We achieve this goal by defining $q_c$ to incorporate human feedback that directly reflects the unknown reward function $R$, in the process of sampling. We do so by first drawing an initial output $x_0$ from a suboptimal LM $\pi_{\theta}$ given the context $c$. Second, we ask humans to rate $x_0$ and provide language feedback $f$ on the $(c,x_0),$ pair. Third, a refinement LM $\pi_{\psi}$ generates a refined output $x_1$ conditioned on $(c,x_0,f)$. The proposal distribution, corresponding to this sampling procedure, can be written down as:

\begin{align}q_c(x_1) = \sum_{f, x_0}\pi_\psi(x_1|x_0,f)p(f|x_0)\pi_\theta(x_0|c). \nonumber
\end{align}

% Since we use a data-generating process that involves humans, we argue that $q_c$ moves us closer to $p^*_c$ than $\pi_{\theta}$. We further assume that $x_0$ is already a good output and that conditioning on feedback further improves it. We further use best-of-N sampling to strengthen this argument. 

Let $x_1^i, \dots, x_1^N$ be $N$ summaries sampled from $q_c(x_1)$. Then, we can approximate the objective in Eq.~\ref{eq:importance_sampling} as: 

\begin{align}
\mathcal{L}_\theta(c)&\approx\sum_{i=1}^N \underbrace{\frac{p^*_c(x_1^i)}{q_c(x_1^i)}}_{=\omega^i} 
\log \pi_\theta(x_1^i|c),
\label{eq:importance_weight}
%\nonumber
% \\
% &=\sum_{i=1}^N \omega^i \log \pi_\theta(x_1^i|c), \\
% &\propto\sum_{i=1}^N \frac{\omega^i}{\sum_{j=1}^N \omega^j} \log \pi_\theta(x_1^i|c) \nonumber
\end{align}

where $\omega^i$ is the importance weight of the $i$-th sample from $q_c$. 
% and in Eq.~\ref{eq:importance_weight} we normalize $\omega^i$'s to sum to 1. 
The importance weight $\omega^i$ is not computable as it is because we do not have access to $q_c$ other than being able to draw samples from it.
We avoid this issue by assuming that $q_c(x_1^i)$ is constant, implying that our samples are all equally good due to the high quality of human feedback. 
We then replace $R(x_1^i|c)$ in the definition of $p_c^*$ by $R(x_1^i|x_0, f, c)$, as the quality is not dependent on the intermediate summary and feedback but can be more easily assessed with these quantities. This allows us to compute the unnormalized $p_c^*$, after which we use self-normalization to finally compute the above loss.
%Second, we replace $R(c,x_1^i)$ with a reward function $R’(c, x^i_0, f^i, x^i_1)$ defined in terms of the original output $x_0^i$ and feedback $f^i$ corresponding to $x^i$ since it is much easier to assess the quality of a summary if you have an original summary with feedback. 

We implement $R$ by conditioning an instruction-finetuned LM on a binary question such as \textit{Does this new text $\left[x_1\right]$ incorporate the feedback $\left[f\right]$ provided on the initial text  $\left[x_0\right]$? Answer Yes or No.}, where the label $y$ is either $y_{\text{good}}$ (`` Yes") or $y_{\text{bad}}$ (`` No"). We use the probability of the positive answer $y_{\text{good}}$ as $R$, i.e. $R(x_1|x_0,f,c) = \frac{p(y_{\text{good}}|\text{prompt})}{p(y_{\text{good}}|\text{prompt}) + p(y_{\text{bad}}|\text{prompt})}$.

Finally, we use an extremely low temperature when computing $p_c^*$, i.e., $\beta \to \infty$. Due to self-normalization, this is equivalent to using only the best summary $x_1^*$ per context $c$ sampled from $q_c$ for computing the loss, resulting in the following, final objective:

%Third, we apply best-of-$N$ sampling against $R'$, i.e. we discard all samples except $x^*_1 = \text{argmax}_{i \in (1,K)} R’(c, x_0^i, f^i, x^i_1)$, which corresponds to setting $\beta\to \infty$, or in other words setting $x_1^*$'s normalized importance weight to 1, and all other outputs normalized importance weights to 0. After these approximations, our objective takes the form 
\begin{align}
\mathcal{L}(\theta) \approx\mathbb{E}_{c\sim p(c)} \log \pi_\theta(x_1^*|c) \label{eq:final_objective}
\end{align}

Our objective of approximating the ground truth distribution $p_c^*(x_1)$, which is proportional to the reward $R$ has clear connections to maximizing reward in RL. However, in RL, the goal is to find the best policy that maximizes the reward, whereas our algorithm results in a distribution of high-quality outputs $x_1$ given a document $c$, which allows us to draw a diverse set of outputs achieving a high reward. The broad diversity of high-quality outputs endows downstream users and systems with more control over which aspects they prefer and want to avoid.
In App.~\ref{app:bayesian_inference_derivation}, we further provide an alternative derivation of ILF that follows variational inference and shows that ILF can also be understood as Bayesian Inference. This process involves updating an LM based on the evidence provided by language feedback. This different lense highlights the correspondence between ILF and RL with Human Feedback ~\citep[][\textit{inter alia}]{ziegler2019fine, stiennon2020learning}, which was previously demonstrated to be equivalent to Bayesian inference~\citep{korbak2022rl}.
%\input{methods_old}

% \section{Experiments}
\section{Can Language Models Use Feedback?}
\label{sec:targeted_word_removal}
For our algorithm to work, LMs must be able to accurately incorporate feedback to generate refinements.
Thus, we first validate the refinement step of our algorithm on a carefully-controlled synthetic task of removing specific offensive words from a given sentence.
We examine how effectively various models incorporate feedback to determine what model to use for refining outputs.

\paragraph{Experimental Setup}
We instruct an LM to refine an automatically-generated sentence with $\leq 10$ offensive words by removing $\leq 3$ specific words (see Appendix \ref{sec:word_removal_example} for a detailed explanation and examples). In this experiment, we generate one output per sample with greedy decoding, i.e., we do not sample with best-of-$N$ . We evaluate how often the generated refinement exactly matches the target sentence, which we automatically generate. For our LMs, we use differently-sized GPT-3 models~\citep{brown2020language} and text-davinci-001, their instruction-finetuned (Feedback Made Easy or FeedME) counterparts~\citep{ouyang2022training,openai_feedme}.\footnote{Via the \href{https://beta.openai.com/}{OpenAI API}.} We report all hyperparameters used in Appendix \ref{app:hp_tuning}.
We report the mean and standard error for all results in our work.

\paragraph{Results}
Table~\ref{tab:targeted_word_removal} shows the results. We observe that only the largest GPT-3 and FeedME models (175B parameters) incorporate feedback in a non-negligible amount of time. Using this insight, we only use the 175B parameter models in the rest of our experiments. Specifically, we use FeedME, because it is an instruction-finetuned model.

\begin{table}[t!]
\resizebox{\columnwidth}{!}{
\begin{tabular}{c c c c c} \toprule
Models  & \makecell{Ada \\ (-)} & \makecell{Babbage \\ (1B)} & \makecell{Curie \\ (6.7B)} & \makecell{Davinci \\ (175B)}  \\
\hline
GPT-3 & $1.2 \pm 0.3$ & $1.7\pm 0.4$ & $8.2 \pm 0.7$ & $38.5 \pm 1.3 $ \\
FeedME & $1.6\pm 0.3$ & $2.2\pm 0.4$ & $6.0\pm 0.6$ & $ 35.8\pm 1.3$ \\

 \bottomrule
\end{tabular}}
\caption{On the task of removing offensive words from a sentence, only large LMs incorporate feedback. We report the percentage of exact string matches with the target.}
\label{tab:targeted_word_removal}
%\vspace{-10px}
\end{table}

\section{Summarization from Language Feedback}
Having established that large LMs can leverage language feedback, we now evaluate our algorithm on the real-world task of text summarization. In \S\ref{sec:data}, we introduce a novel summarization dataset that we use to evaluate our algorithm, in
\S\ref{sec:ranking_refinements}, we explore different methods for ranking refinements and
in \S\ref{sec:comparing_algorithms}, we use the best ranking method to learn from language feedback.

\subsection{Summarization with Language Feedback Dataset}
\label{sec:data}
We evaluate the effectiveness of ILF on the task of text summarization using the TL;DR dataset~\citep{volske-etal-2017-tl}, which consists of Reddit titles, posts, and their corresponding summaries. \citet{stiennon2020learning} adapt this dataset and show that it is a more realistic task for evaluating summarization models compared to the commonly used CNN/DM dataset~\citep{hermann2015teaching}. To ensure the quality of our dataset, we follow the same preprocessing steps as outlined in~\citet{stiennon2020learning} and extract a train dataset with 5000 samples, a development dataset with 200 samples, a validation dataset with 500 samples, and a test dataset with 698 samples\footnote{The train and development datasets are taken from \citet{stiennon2020learning}'s train dataset, and the validation and test set are taken from their test dataset.}. We then hire experienced annotators through Surge AI\footnote{\url{https://surgehq.ai}} to create our language feedback dataset, which we open source along with our code\footnote{Data: \href{https://huggingface.co/datasets/JeremyAlain/SLF5K}{HuggingFace};
Code: \href{https://github.com/JeremyAlain/imitation_learning_from_language_feedback}{Github}}. %make available as supplementary material. 
For each sample, we first generate three summaries for each Reddit post using the instruction-finetuned model text-davinci-001 (FeedME)~\citep{ouyang2022training, openai_feedme}. Two of these summaries are used for a binary comparison, in which annotators indicate their preference. The third summary serves as the initial output for which we solicit language feedback. This feedback should address the single most important shortcoming of the summary and can be related to coverage (how well the summary covers the important information in the post), accuracy (the factual accuracy of the summary), coherence (the coherence of the summary on its own), or other. We do not impose any restrictions on how the feedback should be written. In addition to providing feedback, annotators are also asked to write an ideal summary that is maximally 48 tokens long. The same crowd worker annotates all three tasks for a given sample. Overall the dataset collection and human evaluations cost 40K\$. On selected samples of the binary comparison task, we achieve an author-annotator agreement of $81.0\%$ and annotator-annotator agreement of $70.0\%$. The human summaries we collect are of excellent quality, as demonstrated in a human evaluation, where we compare our human-written summaries to the ones automatically extracted from Reddit \citep{volske-etal-2017-tl} (also used as baselines in \citet{stiennon2020learning,scheurer2022training}). We find that our human-written summaries are preferred $72.0 \pm 3.2 \%$ of the time, making them a much stronger baseline.

\begin{table}[t!]
\resizebox{\columnwidth}{!}{
\begin{tabular}{c c c} \toprule
 & Scoring Function &  \makecell{Win Rate in \% vs. \\ Random Selection} \\
\toprule
\makecell{Task Specific \\ Heuristic} & Max Length & $65.0 \pm  2.7$\\ 
\midrule
\midrule
\multirow{7}{*}{Zero-Shot} & 
Embedding Similarity   &  
$48.3 \pm 3.0$\\
& InstructRM Prompt 1 & $55.0 \pm 3.0$ \\
& InstructRM Prompt 2 & $58.0 \pm 2.9$ \\
& InstructRM Prompt 3 & $56.5 \pm 2.9$ \\
& InstructRM Prompt 4 & $55.8 \pm 2.8$ \\
& InstructRM Prompt 5 & $50.0 \pm 3.0$ \\
& \textbf{InstructRM Ensemble} & \textbf{56.0} $\pm$ \textbf{3.0}\\
\bottomrule
\end{tabular}}
\caption{We compare various ranking methods for selecting refinements using a human evaluation. InstructRM Ensemble performs best and is used throughout our paper.}
\label{tab:scoring_function_results} 
%\vspace{-12px}
\end{table}

\subsection{Comparing Refinement Ranking Methods}
\label{sec:ranking_refinements}
\paragraph{Generating Refinements}
We condition FeedME on the initial summaries of our train dataset (generated with FeedME) and the human-written feedback and generate 5 refinements $x_1^1, ...,x_1^5$ using the instructions in App.~\ref{app:summarization_prompts}.

%We use initial summaries (generated with FeedME) and the human-written feedback of our train dataset with 5000 samples We condition FeedME-175 on the initial summaries and the feedback and generate 5 refinements $x_1^1, ...,x_1^5$ using the instructions in App. \ref{app:summarization_prompts}.

\paragraph{Scoring Refinements with InstructRM} \label{sec:scoring_functions} We chose a refinement with a scoring function $R$ that scores refinements for how effectively they incorporate feedback. 
%For $R$ we use the instruction finetuned LM to evaluate its own output and call this method \textit{Instruct Reward Model (InstructRM)}. 
For $R$ we use the instruction-finetuned LM FeedME and ask it whether a refinement is better than the initial summary  (see \S\ref{sec:methods} for more details). We then evaluate the probability that the refinement incorporates language feedback on the initial summary and is accordingly a high-quality summary, i.e., $p(y_{\text{good}}|\text{prompt})$.
%e.g., $p(\text{improved}) = \frac{{p(\textit{" Yes"} | \text{prompt})}}{p(\textit{" Yes"}|\text{prompt}) + (p(\textit{" No"}|\text{prompt})}$.
LMs are sensitive to the exact prompt used~\citep{perez2021true,lu2021fantastically}, so we write 5 different prompts (see App.~\ref{app:instruct_rm_prompts})
and select the refinement with the highest average $p(y_{\text{good}}|\text{prompt})$ and call this method InstructRM Ensemble.

\paragraph{Scoring Refinements with Embedding Similarity}
Previous work \citep{scheurer2022training} use a contrastive pre-trained text-embedding function \citep{neelakantan2022text} to embed the feedback $f$ and refinements $x_1^1, ...,x_1^5$ and select the refinement with the highest cosine similarity to the feedback. They use this scoring function because feedback would often describe what the ideal text should look like. This method is less general because it assumes that good refinements are semantically similar to the feedback, which is not necessarily the case for all tasks or forms of feedback.

%Previous work \citep{scheurer2022training} use a contrastive pre-trained text-embedding function \citep{neelakantan2022text} to embed the feedback $f$ and refinements $x_1^1, ...,x_1^5$. They then choose the refinement with the highest cosine similarity with the feedback. They chose this scoring function because feedback would often describe what the ideal or improved text would look like. This method is less general because it assumes that good refinements are semantically similar to the feedback, which is not necessarily the case for all tasks or forms of feedback.

%In reality, however, this scoring function has clear limitations that become apparent when going beyond simple feedback that mentions missing information. It is unclear how an embedding similarity could handle feedback that talks about removing information or making stylistic comments etc. We conducted an analysis where we compared our InstructRM scoring function, with the embedding similarity scoring function. 

\paragraph{Results}
\label{sec:scoring_function_results}
We now evaluate the above ranking methods on the development dataset by calculating the fraction of times the refinement selected by a method is better than a randomly-selected refinement (``win rate''), according to a ranking given by human evaluators (see App.~\ref{app:ranking} for more details). The results, shown in Table~\ref{tab:scoring_function_results}, show that the embedding similarity selection does not outperform random selection, while most (4/5) InstructRM prompts do. While the embedding similarity worked well in previous work~\citep{scheurer2022training}, it does not perform well on our dataset. We believe this is because the feedback we collect, written by many annotators, is much more diverse, while in \citet{scheurer2022training}, the authors wrote the feedback themselves. InstructRM Ensemble has a win rate of $56.0 \pm 3.0 \%$ against random selection, demonstrating that an LM can evaluate its own output to some extent. Based on these results, we recommend using the InstructRM Ensemble approach, as it performs well and is less sensitive to the particular prompt.
%As a result, the method does not rely on a development dataset to select a specific prompt.% determine if a prompt is outperforming random selection.
Throughout our paper, we use InstructRM Ensemble as our scoring function to select refinements and refer to our method of generating and selecting refinements as \textit{Refinement with Feedback + Best of N}.
% We evaluate the ranking quality by comparing against a human-evaluated ranking of the selected refinement, with the rank of a randomly selected refinement.

\begin{figure}[t!]
   \begin{center} 
    \small{
    \scalebox{0.4}{\cstar{150}{199}{157}} ILF + OPT-RM (best-of-64)\quad \\
    \cblock{191}{134}{173} OPT-RM best-of-64 FeedME \quad \cblock{160}{125}{108} FeedME \quad\\
    \cblock{206}{119}{107} ILF: Finetuned on Refinements \quad
   \cblock{243}{192}{125} Finetuned on Initial Summaries \quad
   \cblock{125}{194}{209} Finetuned on Human Summaries 
   }
    \end{center}
    \includegraphics[scale=0.48]{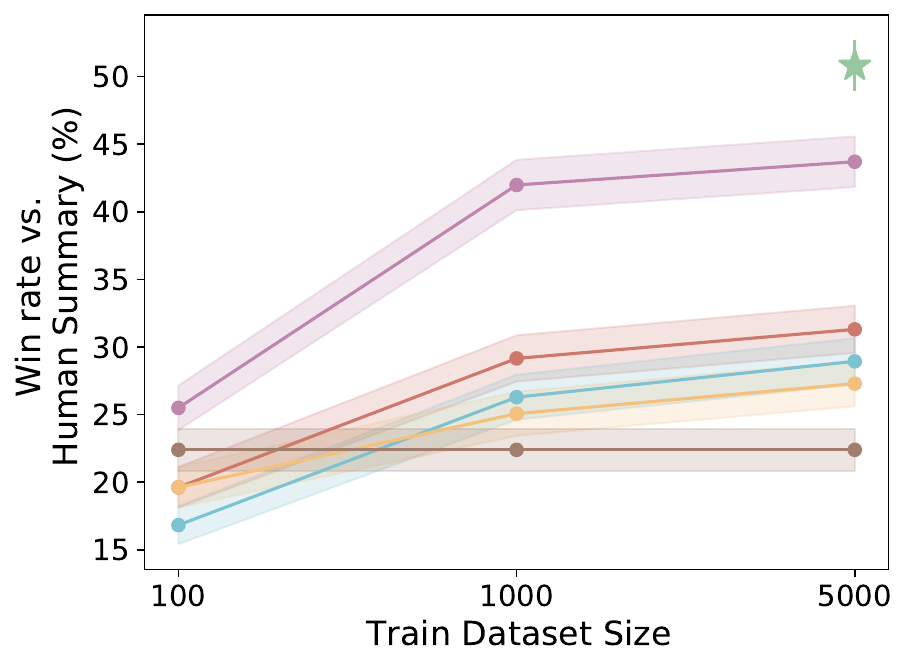}
\caption{How often human evaluators prefer summaries from ILF, OPT-RM best-of-64 FeedME, ILF + OPT-RM (best-of-64), finetuning baselines and FeedME to human summaries. ILF + OPT-RM (best-of-64) generates summaries of a similar quality to human summaries.}
    \label{fig:finetuned_methods_comparison}
\end{figure}

\subsection{Comparing Feedback Learning Algorithms}
\label{sec:comparing_algorithms}
In this section, we compare various algorithms for learning from language feedback, binary feedback, and normal supervised finetuning. We present an overview of each method and then provide the results of our evaluations. 

\subsubsection{Methods}
\label{sec:finetuning_refinements}
\paragraph{Finetuning on Refinements (ILF)}
%Our Expert Iteration algorithm is designed to incorporate human preferences into the training of language models. This is achieved through an iterative process that leverages feedback in the form of language.
For this evaluation, we use a single iteration of ILF to learn from language feedback.
We finetune GPT3-175B (davinci)~\citep{brown2020language}\footnote{FeedME cannot be finetuned via OpenAI's API.} to 
maximize the log-likelihood of the refinement given the input prompt (consisting of the Reddit title, and post), i.e., $\log p(x_1 | \text{prompt})$, using the refinements generated with Refinement with Feedback + Best of N. %maximize the log-probability of the refinement given the initial summary $\log p(x_1 | x_0)$ using the \textsc{Refinement with Feedback + Best of N} dataset.
For all our finetuning methods we add $\lambda \log p(\text{prompt})$ to the loss \citep{radford2018improving,openai_documentation}, which maximizes the log-probability of the prompt. The prompt-loss weight $\lambda \in [0, 1]$ is chosen on our development dataset (see paragraph \textit{Finetuning on Human Summaries}). The selected hyperparameters are detailed in App.~\ref{app:hp_tuning} and the finetuning prompts in App.~\ref{app:finetuning_prompts}.

%The full prompt templates that we used for finetuning can be viewed in Appendix~\ref{app:finetuning_prompts}.
%We do extensive hyperparameter tuning, (see paragraph \textit{Finetuning on Human Summaries}) and  provide additional details in Appendix \ref{app:hp_tuning} for the selected hyperparameters. 
\paragraph{Finetuning on Human Summaries}
\label{sec:finetuning_human_summaries_hp}
Here we finetune GPT3-175B on the dataset of human-written summaries $x_{\text{human}}$, with the objective of maximizing the log-probability of human summaries given the input prompt (consisting of the Reddit title and post) with the additional loss term, i.e. $\log p(x_{\text{human}} | \text{prompt}) + \lambda \log p(\text{prompt})$.
To ensure the best performance of our finetuned models, we conduct 
 thorough hyperparameter tuning on the human-written summary datasets of various sizes (100, 1K, 5K). The hyperparameters optimized include the number of training epochs, the prompt loss weight $\lambda$, and the learning rate multiplier, as detailed in the OpenAI documentation \citep{openai_documentation}. We use the perplexity of the predicted summaries on the development dataset to select the most effective hyperparameters. The selected hyperparameters are applied to all datasets, i.e., finetuning on refinements, initial summaries, and human-written summaries, with the same sample size.  More details on hyperparameter tuning can be found in Appendix~\ref{app:hp_tuning}.

\paragraph{Finetuning on Initial Summaries}
We finetune GPT3-175B on the dataset of initial summaries (generated by FeedME). The objective is to maximize the log probability of the initial summary given the prompt (consisting of the Reddit title and post) with the additional loss term i.e. $\log p(x_0 | \text{prompt}) + \lambda \log p(\text{prompt})$. Details on hyperparameter tuning can be found in the paragraph \textit{Finetuning on Human Summaries} and Appendix~\ref{app:hp_tuning}.

 %To select the most effective model, we evaluate all finetuned models on the development dataset and use the perplexity of the predicted summaries as our selection criteria. Given that the human summary dataset, the initial summary dataset, and the refinement dataset only differ in the actual summaries, we apply the same selected hyperparameters to all datasets with the same sample size. We refer to Appendix~\ref{app:hp_tuning} for more details on hyperparameter tuning.

%We do extensive hyperparameter tuning on the human summaries, for each of the datasets splits with 100, 1K and 5K samples. We optimize the number of training epochs, the prompt loss weight $\lambda$, and the learning rate multiplier\footnote{See OpenAI's documentation \citep{openai_documentation} for more details on the hyperparameters}. We evaluate all finetuned models on the development dataset and use the perplexity of the predicted summaries as selection criteria. Since the human summary dataset, the initial summary dataset and the refinement dataset only differ in the actual summaries, we use the same selected hyperparameters for all datasets with the same amount of samples. 

% The prompt is however much longer than the completion and accordingly has a much higher influence on the loss. We thus introduce a hyperparameter called the prompt loss weight $\lambda \in [0...1]$, which controls the weight of the loss for the prompt tokens. 

\paragraph{Learning from Binary Feedback: Best-of-$N$}
\label{sec:binary_feedback}
We compare ILF against binary feedback as a baseline, the standard approach for learning from feedback. One way of learning from binary feedback is to train a reward model and use it to do best-of-$N$ sampling. We use best-of-N because it is often competitive with RL from human feedback~\citep{nakano2021webgpt}, a highly effective but more sophisticated approach~\cite{stiennon2020learning,ouyang2022training}.
To train the RM, we finetune OPT-13B (OPT-RM) \citep{zhang2022opt} to classify whether a summary $x_0$ is high quality or not.
To do so, we use the instruction \textit{Is the above an excellent summary of the given text? An excellent summary is coherent, accurate, concise, and detailed. Answer with Yes or No.}, where the label $y$ is either $y_{\text{good}}$ (`` Yes") or $y_{\text{bad}}$ (`` No"). Given human labels on which of two summaries is preferred, we label the preferred summary with $y_{\text{good}}$ and the other summary with $y_{\text{bad}}$.
We then finetune the LM to maximize $\log p(y | x_0) + \lambda \log p(x_0)$, where $\lambda \in [0,1]$, chosen using the development dataset, and $y \in \{y_{\text{good}}, y_{\text{bad}}\}$.
Using the finetuned LM, we evaluate a given summary by computing $p(y_{\text{good}}|x_0)$ and select the summary with the higher probability. We find that this approach leads to more accurate RMs than other RM training methods, such as the commonly used method from \citet{stiennon2020learning}; see Appendix~\ref{app:reward_model} for comparisons and Appendix~\ref{app:rm_prompts} for the used prompts. We perform Bayesian hyperparameter optimization for OPT-RM and sweep over the learning rate, batch size, and prompt-loss weight $\lambda$, using classification accuracy on the development dataset as the selection criteria (see Appendix~\ref{app:hp_tuning} for more details).

%use the above approach throughout our paper since we found that it leads to more accurate RMs than other RM training methods, such as the commonly used method from \citet{stiennon2020learning}; see Appendix~\ref{app:reward_model} for comparisons and Appendix~\ref{app:rm_prompts} for the prompt templates used. We perform a Bayesian hyperparameter optimization for OPT-RM and sweep over the learning rate, the batch size, and the prompt-loss weight. We use classification accuracy on the development dataset as a selection criterion (see Appendix~\ref{app:hp_tuning} for more details).

%$p(\text{summary is high-quality}) = \frac{p(y_{\text{good}}|x_0)}{p(y_{\text{good}}|x_0) + p(y_{\text{bad}}|x_0)}$
%We then pick the one with the higher probability for \textit{is high-quality summary}. 

\paragraph{ILF + Learning from Binary Feedback}
\label{sec:ILF_OPT}
As a final step, we combine ILF and learning from binary feedback, by first finetuning GPT3-175B on the refinements as described in the paragraph finetuning on refinements (ILF). We then train the reward model, OPT-RM, and use it to perform best-of-$N$ sampling, as outlined in the paragraph on learning from binary feedback. At test time, we generate 64 summaries with our finetuned model and rank them based on their 
probability of being a high-quality summary, $p_\text{norm}(y_{\text{good}}|x_0)$, using OPT-RM. The summary with the highest normalized probability is then selected.
%We begin by finetuning GPT-3 \textit{davinci} on refinements as described in the previous paragraph \textit{Finetuning on Refinements (ILF)}. We then follow the process outlined in the paragraph on learning from binary feedback to train the reward model, OPT-RM, and use it to perform best-of-$N$ sampling. At test time, we generate $N$ summaries using our finetuned model and use OPT-RM to rank them based on their normalized probability of being a high-quality summary, $p_\text{norm}(y_{\text{good}}|x_0)$. The summary with the highest normalized probability is then selected.

%Finally, we introduce the combination of ILF and Learning from Binary Feedback. First finetung \textit{davinci} on refinements as described in Finetuning on Refinements (ILF). Then train OPT-RM as described in Learning from Binary FeedbacK: Best-of-$N$. At test time we sample $N$ summaries from our finetuned model and use OPT-RM to rank them. We then select the summary with the highest normalized probability $p_\text{norm}(y_{\text{good}}|x_0)$.

\subsubsection{Evaluation}
\label{sec:evaluation_results}
We evaluate the effectiveness of our learning algorithm, by comparing it to human written reference summaries, several finetuning baselines, and OPT-RM on the task of text summarization using 100, 1K, and 5K train samples. Using a test dataset of 698 samples, we generate a summary for each method and evaluate them with human evaluators who rank them based on quality, using a standard ranking scheme that allows for ties between summaries (see App.~\ref{app:hp_tuning} for more details). Based on the rankings, we calculate the fraction of times each method's sampled summary outperforms the human-written reference summary, referred to as the ``win rate''. We sample summaries up to 48 tokens in length (as in \citet{stiennon2020learning}) using nucleus sampling \citep{holtzman2019curious} with $p=0.95$ and temperature $t=1.0$ (see App.~\ref{app:hp_tuning} for further details on hyperparameters and postprocessing). We use best-of-64 sampling with summaries sampled from FeedME for learning from binary feedback.

\subsubsection{Results}
\label{sec:main_results}
\begin{figure}[t!]
\begin{minipage}[t]{.45\textwidth}
    \centering
\includegraphics[scale=0.48]{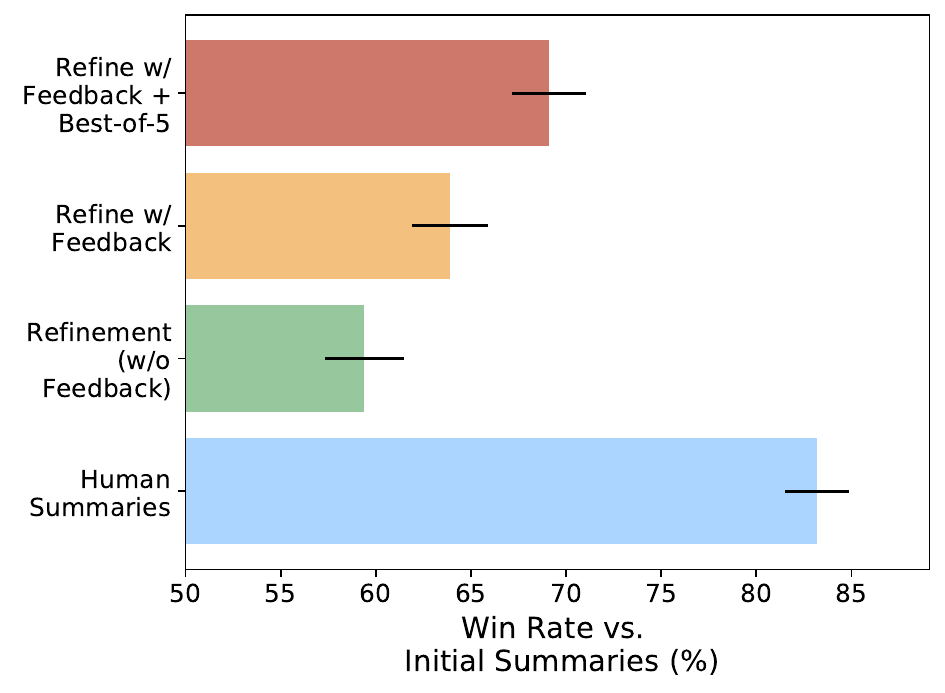}
\end{minipage}
\hfill
\begin{minipage}[t]{.45\textwidth}
     \centering
\includegraphics[scale=0.48]{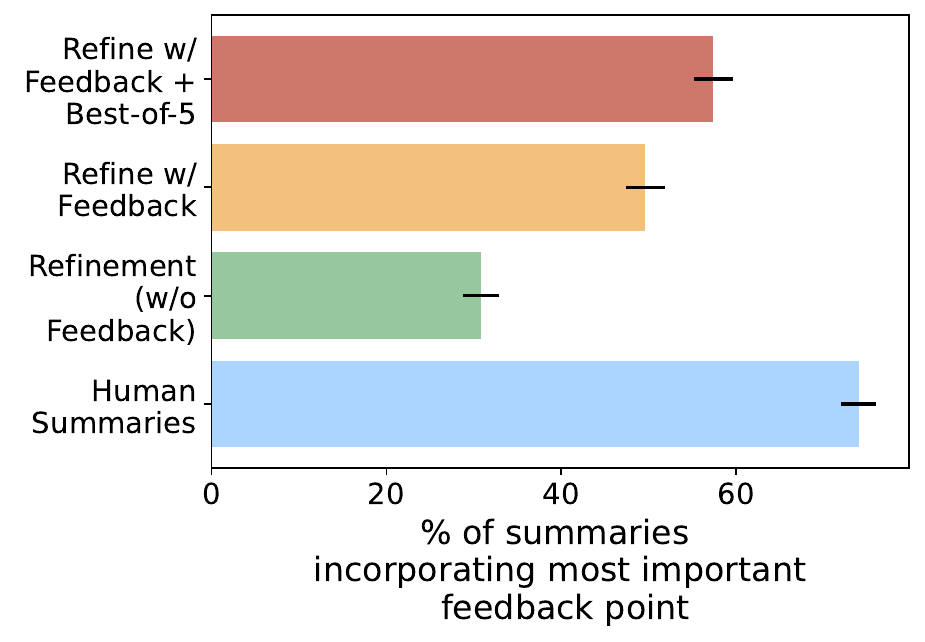}
\end{minipage}
\caption{\textbf{Top}: Human evaluators prefer summaries from
all refinement methods to the initial summaries (FeedME). Refine with Feedback + best-of-5 is rated highest. \textbf{Bottom}: Refine with Feedback + best-of-5 generally does incorporate the most important feedback point.}
\label{fig:win_rates_and_incorporating_feedback}
\vspace{-10px}
\end{figure}

Our results, shown in Fig.~\ref{fig:finetuned_methods_comparison}, demonstrate that finetuning on refinements (ILF) outperforms all other finetuning methods\footnote{Finetuning on 100 refinements is tied with finetuning on 100 initial summaries.}), including sampling from FeedME, with a win rate against human summaries of $31.3 \pm 1.7 \%$ (for finetuning on 5K samples), while the other methods achieve win rates of $27.3 \pm 1.7 \%$ (finetuning on initial summaries), $28.9 \pm 1.7 \%$ (finetuning on human summaries), and $22.5 \pm 1.6 \%$ (FeedME). It is surprising that ILF outperforms finetuning on human summarise across all sample sizes, despite human-written summaries generally being of higher quality (see Fig.~\ref{fig:win_rates_and_incorporating_feedback}, top). Further evaluation (see App. Fig.~\ref{fig:finetuned_methods_loss_scaling}) shows that the model finetuned on 1K refinements (ILF) exhibits significantly lower loss when evaluated on the validation dataset of refinements compared to the model finetuned on human summaries when evaluated on the validation dataset of human summaries, suggesting that the model is more adept at approximating the distribution of refinements. Additionally, when evaluating GPT3-175B on the summaries of 1K samples from various train datasets, we observe significantly lower loss on the refinement dataset than on the dataset of human summaries (see Table.~\ref{tab:Kl_distance_of_finetuned_models}). Overall, these results demonstrate the effectiveness of our proposed ILF approach in accurately incorporating feedback and improving model performance, even outperforming finetuning on human summaries.

\citep{scheurer2022training} found that ILF with 100 feedback samples outperformed FeedME, while here we find it underperforms FeedME with 100 feedback samples.
Prior work uses author-written feedback that often conveys what the refinement should include, while our work includes more varied, crowdsourced feedback.
As a result, we observe that embedding similarity does not properly rank refinements on our human feedback dataset (Table~\ref{tab:scoring_function_results}), and we believe the difference in feedback may be a significant source of differences in results in this section as well; see Appendix~\ref{app:comparison_scheurer_22} for more discussion. 

Our results demonstrate that using OPT-RM for best-of-64 sampling on FeedME summaries outperforms all finetuning methods and sampling approaches across all sample sizes. The improved performance of OPT-RM best-of-64 FeedME comes at the cost of added inference time for best-of-$N$ sampling. Combining ILF and learning from binary feedback (ILF + OPT-RM (best-of-64)) achieves human-level summarization performance with a win rate of $50.8 \pm 1.9 \%$ using 5K samples for training. This suggests that both methods independently learn valuable information about human preferences that can be cumulative when used together. It should be noted that the result for ILF + OPT-RM (best-of-64) is obtained through a separate human evaluation with different comparison summaries (see App. Fig.~\ref{fig:additional_results}), and was added to Fig.~\ref{fig:finetuned_methods_comparison} for reference. 
%The results of the evaluation shown in Fig.~\ref{fig:additional_results}, demonstrate that OPT-RM best-of-64 FeedME achieves a win rate of $45.1 \pm 1.9 \%$, while here (see Fig.~\ref{fig:finetuned_methods_comparison}) it achieves a comparable win rate of $43.7 \pm 1.9 \%$.
In App.~\ref{app:ilf_multiple_iteratins}, we present some initial, promising results for multiple iterations of ILF. These results suggest that the method is effective, but further experimentation is necessary to understand it better.

\subsection{Does Language Feedback Improve Refinements?}
The improvements from ILF suggest that the refinements used for finetuning are high-quality, so here we investigate whether language feedback is responsible for the high quality.
% Our algorithm relies on having high-quality refinements to finetune on, so here, we investigate which aspects of our algorithm are responsible for improving the refinement quality.
%We now aim to examine the importance of various aspects of our algorithm for generating high-quality refinements (before finetuning).
To do so, we have human evaluators rank Refinement with Feedback + Best of N summaries against summaries from several other methods, similar to \S\ref{sec:ranking_refinements}.
We use the human ranking to compute a win rate between each method and the initial summary.
We compare against Refinement with Feedback, which \textit{randomly} chooses a refinement $ \in {x_1^1, \dots, x_1^5}$.
This ablation helps to evaluate the importance of choosing a refinement with our scoring function $R$, i.e., InstructRM Ensemble.
We also evaluate Refinement without Feedback, which instructs the LM to refine the initial summary but without feedback.
This ablation helps to evaluate the importance of using language feedback.
Lastly, we evaluate Human Summaries and Initial Summaries i.e., the initial summary $x_0$ generated by FeedME. We evaluate all methods on the validation dataset.

\paragraph{Results.}
Fig.~\ref{fig:win_rates_and_incorporating_feedback} (top) shows the win rates of summaries from various methods against initial summaries.
Surprisingly, instructing a model to improve its output without feedback already leads to a significant improvement (win rate of $59.4 \pm 2.1 \%$ over the initial summaries).
Refinements with Feedback achieve an improved win rate of $63.9 \pm 2.0 \%$, showing that language feedback is useful for improving refinement quality.
Refinement with Feedback + Best of N achieves an even better win rate of $69.1 \pm 1.9 \%$, highlighting that Best-of-N with the InstructRM Ensemble further improves the refinements.
Overall, language feedback is important for high-quality refinements, especially when using Best-of-N sampling.

\subsection{Do Refinements Incorporate the Feedback?}
To determine whether refinements are of higher quality due to incorporating feedback rather than improving the summary in other ways, we conduct a study on the validation dataset in which crowd workers evaluate how often the most important point of the feedback is incorporated in the refinements produced by various methods. As shown in Fig.~\ref{fig:win_rates_and_incorporating_feedback}, bottom, our method Refinement with Feedback + Best of N incorporates the most important point in the feedback most frequently ($57.4 \pm 2.2 \%$ often). Refinement with Feedback incorporates feedback $49.6 \pm 2.2\%$ of the time, showing that Best-of-N sampling improves how often the feedback is incorporated. For reference, Refinement without Feedback fixes the most important point in the feedback $30.8 \pm 2.1\%$ of the time, despite the model not receiving the language feedback. Human Summaries address the most important point in the feedback $74.0 \pm 1.9 \%$ of the time when writing the summary from scratch despite not receiving the feedback explicitly. Our results suggest that refinements are high-quality in part because they incorporate the most important point in the feedback.

%Here, we examine whether refinements are of higher quality because they incorporate the feedback, rather than improving the summary in other ways.
%To do so, we have human evaluators evaluate how often the most important point in the feedback is incorporated in the refinement, for various refinement methods.
%Fig.~\ref{fig:win_rates_and_incorporating_feedback} (right) shows that our method \textsc{Refinement with Feedback + Best of N} incorporates the most important feedback point most of the time ($57.4 \pm 2.2 \%$ often). \textsc{Refinement with Feedback} incorporates feedback $49.6 \pm 2.2\%$ of the time, showing that Best-of-N sampling improves how often the feedback is incorporated.
%For reference, \textsc{Refinement without feedback} fixes the most important feedback point $30.8 \pm 2.1\%$ of the time, despite the model not receiving the language feedback.
%The \textsc{Human Summaries} have addressed the most important feedback point $74.0 \pm 1.9 \%$ of the time when writing the summary from scratch and despite not receiving the feedback explicitly.
% and human summaries act as counterfactuals, i.e., they never observed feedback. They measure, however, how often a direct refinement without feedback would, by chance, incorporate feedback, and how much an ideal, human-written summary would address feedback on a fictitious summary. The former incorporates feedback $30.8 \pm \%$ of the time, and the latter $74.0 \pm 1.9 \%$ of the time.

\subsection{Which Finetuning Dataset Changes Models Most?}
\label{sec:results_distribution}
Here, we aim to understand how the summaries used for finetuning influence how much the model changes after finetuning.
\citet{gao2022scaling} find that models optimized with binary human feedback are more likely to learn undesirable behaviors when their output distribution deviates more from the initial, pretrained LM.
It is unclear whether these findings apply to models trained with language feedback, but
we take a preliminary step in this direction for understanding language feedback-trained models. % understanding language feedback-trained models.
In particular, we measure the (reverse) KL divergence \citep[following][]{gao2022scaling} between an ILF-finetuned model and the pretrained LM before ILF-training, $D_{\text{KL}}(\text{finetuned}|\text{GPT3-175B})$, by unconditionally sampling from the finetuned model and evaluating the log-likelihood of the generated text with \text{GPT3-175B}.
We also report the forward KL divergence, $D_{\text{KL}}(\text{GPT3-175B} | \text{finetuned})$. 
For reference, we evaluate both of the above for models finetuned on the initial summaries and on human summaries.

\paragraph{Results.}
Finetuning on refinements (ILF) shows the largest KL divergence (in both directions), followed by finetuning on human summaries, and then followed by finetuning on initial summaries; see App. Table~\ref{tab:Kl_distance_of_finetuned_models} for the exact numbers.
We find it surprising that finetuning on refinements results in higher KL divergences than finetuning on human summaries; we expected the refinements to be closer to the model's initial output distribution, relative to human summaries, therefore causing the finetuned model to undergo less change. The larger KL divergence with ILF may be partly responsible for the larger gains in human evaluations observed in Fig.~\ref{fig:finetuned_methods_comparison}.

%First, we calculate the negative log-likelihood of the \textit{davinci} model on the train dataset with 1K samples of the initial summaries, the refinements, and the human summaries. We find that the human summaries have the highest negative log-likelihood, followed by refinements and then initial summaries. This suggests that the human summaries are the farthest in terms of KL distance from the distribution learned by \textit{davinci}. This could potentially explain why it is more challenging to learn the human summary distribution, especially in the low data regime.

%We further calculate the forward and backward KL distances between the various finetuned models and \textit{davinci}. The forward KL distance is calculated by sampling $64$ tokens from \textit{davinci} and evaluating the generated text with the finetuned models, while the backward KL distance is calculated by sampling $64$ tokens from a finetuned model and evaluating \textit{davinci} on this text. The results, shown in Tablele~\ref{tab:Kl_distance_of_finetuned_models}, reveal that, in both the forward and backward KL divergences, the model finetuned on refinements is the furthest from \textit{davinci}, followed by finetuning on initial summaries and then human summaries. This suggests that the model finetuned on refinements is the most optimized . Lastly, we provide a PoS analysis of the 1K datasets of those summaries in Appendix \ref{app:pos}.
\section{Related Work}
Our work builds upon our previous report \citep{scheurer2022training}, which showed that large LMs can refine outputs with language feedback. There, we introduce the same three-step algorithm that ILF builds upon, with the key difference that here we use an LM, i.e., InstructRM Ensemble, to evaluate whether a refinement incorporates feedback, whereas in \citet{scheurer2022training} we use a contrastive pre-trained text-embedding function \citep{neelakantan2022text}. InstructRM Ensemble is more general than this Embedding Similarity since it does not assume semantic similarity of the refinements to the feedback. Another difference is that ILF is an iterative, refine-and-finetune algorithm, which can be understood as Bayesian Inference corresponding to RL with Human Feedback. In addition, here we conduct different and more extensive experiments than in \citet{scheurer2022training} and  use human annotators. In particular, we show that ILF outperforms finetuning on human summaries and that combining ILF with learning from binary feedback achieves roughly human-level summarization performance. For a more detailed comparison to \citet{scheurer2022training} we refer to App.~\ref{app:comparison_scheurer_22}.

Subsequent work to ours suggests several ways to improve upon our approach.
\citet{saunders2022self} show that LMs themselves write high-quality feedback on LM outputs.
\citet{bai2022constitutional} then train a dialog assistant using ILF to learn from LM-written language feedback, eliminating the cost and effort of collecting human feedback.
\citet{liu2022improving,schick2022peer} train LMs to refine outputs based on feedback (without finetuning on the refinements), an approach that improves results when incorporated into ILF, as shown in subsequent work to ours \citep{shi2022life}.

Other work aims to use language in other ways than we do.
Some work investigates using explanations for \textit{gold labeled outputs} to \textit{classification tasks}, while our work addresses the more general text generation setting which classification tasks can be formulated as~\cite{Radford2019LanguageMA,raffel2020exploring,brown2020language}.
Explanations describe why a labeled output is correct, while feedback describes how to improve a candidate's output.
Prior work explores ways of using explanations to train text classification models, with mixed results~\citep[][\textit{inter alia}]{camburu2018snli,stacey2021natural,pruthi2021evaluating,wiegreffe-etal-2021-measuring,hase2021can, lampinen2022can}.
A few prior works also learn from language feedback for the purpose of ranking candidate outputs rather than generating outputs~\citep[]{weston2016dialog, li2016dialogue, hancock2019learning,li2022using,xu2022learning}.
\citet{matiana2021cut} learn text embeddings of language feedback, where improvements could benefit the refinement-scoring step of our algorithm. Language has also been used for various purposes in RL settings as well, as discussed in App.~\ref{app:releated_works_rl}.
% Other work learns to rank~\citep[]{weston2016dialog, li2016dialogue, hancock2019learning}, or generate \citep{li2022using} textual feedback, which serves as supervision signal to improve overall performance on question answering tasks. Futhermore, \citep[]{weston2016dialog, li2016dialogue, hancock2019learning} use ranking algorithms to select answers, and while \citep{li2022using} does learn to generate feedback, their final model is also used to rank answers. We on the other hand do not predict feedback, but directly leverage it to improve open ended text genertation. Lastly, \citet{matiana2021cut} learn text embeddings of language feedback, where improvements could benefit the refinement-scoring step of our algorithm.

Several other works draw connections between Bayesian Inference and learning algorithms for LMs.
\citet{korbak2022rl} show that KL-regularised RL is equivalent to variational inference: approximating a Bayesian posterior which specifies how to update a prior LM to conform with evidence provided by a reward function. \citet{dohan2022language} further argues that the process of generating output through multiple rounds of interaction between prompted LMs and other agents (e.g. humans providing language feedback) can be seen as executing probabilistic programs.
\section{Conclusion}
In this work, we propose Imitation learning from Language Feedback (ILF), an iterative algorithm for training LMs to behave in line with human preferences, by learning from language feedback. We validate our approach on a carefully-controlled word-removal task, showing that only large LMs (175B parameters) accurately incorporate feedback. Using this insight, we then test our algorithm on the real-world task of text summarization. Combining ILF and learning from binary feedback brought a GPT-3 model to roughly human-level summarization ability. ILF on its own outperformed finetuning on human summaries, despite human summaries being of higher quality, suggesting that the model is better at approximating the distribution of refinements. %Our analysis further demonstrated that refinements generated with LMs typically incorporate feedback, especially when using an LM to select refinements.
% Initial results suggest that multiple iterations of ILF can incorporate human preferences even better.
%Language feedback is a natural form of communicating with models, making ILF a promising algorithm to align LM outputs with human preferences.
Our work opens up many avenues for future work, from improving algorithms for learning from language to tackling settings where it is hard to learn from sparse or binary feedback.

\section{Acknowledgements}
We are grateful to Nat McAleese, Geoffrey Irving, Jeff Wu, Jan Leike, Cathy Yeh, William Saunders, Jonathan Ward, Sam Bowman, Daniel Ziegler, Seraphina Nix, Quintin Pope,  Kay Kozaronek, Peter Hase, Asa Cooper Stickland, Jacob Pfau, David Lindner,  Lennart Heim, Nitarshan Rajkumar, Kath Lumpante, Pablo Morena, Edwin Chen, Scott Heiner, and David Dohan for helpful conversations and feedback.
Jérémy Scheurer and Jun Shern Chan thank Open Philanthropy for funding that enabled this research.
Ethan Perez thanks the National Science Foundation and Open Philanthropy for fellowship support.
Jon Ander Campos is supported by a doctoral grant from the Spanish MECD.
Angelica Chen and Kyunghyun Cho are supported by the NYU Center for Data Science National Science Foundation (Award 1922658).
KC was supported by 42dot, Hyundai Motor Company (under the project Uncertainty in
Neural Sequence Modeling), Samsung Advanced Institute of Technology (under the project Next
Generation Deep Learning: From Pattern Recognition to AI), and NSF Award 1922658 NRT-HDR:
FUTURE Foundations, Translation, and Responsibility for Data Science
We also thank OpenAI for providing access and credits to their models via the API Academic Access Program.
\newpage

\bibliography{references}
\bibliographystyle{icml2023}

%%%%%%%%%%%%%%%%%%%%%%%%%%%%%%%%%%%%%%%%%%%%%%%%%%%%%%%%%%%%%%%%%%%%%%%%%%%%%%%
%%%%%%%%%%%%%%%%%%%%%%%%%%%%%%%%%%%%%%%%%%%%%%%%%%%%%%%%%%%%%%%%%%%%%%%%%%%%%%%
% APPENDIX
%%%%%%%%%%%%%%%%%%%%%%%%%%%%%%%%%%%%%%%%%%%%%%%%%%%%%%%%%%%%%%%%%%%%%%%%%%%%%%%
%%%%%%%%%%%%%%%%%%%%%%%%%%%%%%%%%%%%%%%%%%%%%%%%%%%%%%%%%%%%%%%%%%%%%%%%%%%%%%%
\newpage

\appendix
\onecolumn
\section{Additional derivations}
\label{app:additional_derivation}

\subsection{Imitation Learning from Language Feedback as Bayesian Inference}
\label{app:bayesian_inference_derivation}

\subparagraph{Language Feedback as Variational Inference}

\begin{figure}[b]
\centering
\begin{minipage}{0.48\textwidth}
   % \vspace{-12px}
    \begin{subfigure}{0.2\textwidth}
           \scalebox{0.9}{\begin{tikzpicture}
  % Define nodes
  \node[obs]           (c) {$c$};
  \node[latent, right=0.6cm of c]       (x_1) {$x_1$};
  \node[latent, right=0.6cm of x_1]        (I) {$\mathcal{I}$};

  % Connect the nodes
  \edge {c} {x_1} ; %
  \edge {x_1} {I} ; %
  \draw [->] (c) to [out=40,in=140] (I);

\end{tikzpicture}}
    \end{subfigure}
\hspace{0.25\textwidth}%
    \begin{subfigure}{0.2\textwidth}
        \scalebox{0.9}{\begin{tikzpicture}
  % Define nodes
  \node[obs]           (c) {$c$};
  \node[latent, right=0.6cm of c]       (x_0) {$x_0$};
  \node[latent, right=0.6cm of x_0]     (f) {$f$};
  \node[latent, right=0.6cm of f]       (x_1) {$x_1$};
  \node[obs, right=0.6cm of x_1]        (I) {$\mathcal{I}$};

  % Connect the nodes
  \edge {c} {x_0} ; %
  \edge {x_0} {f} ; %
  \edge {f} {x_1} ; %
  \edge {x_1} {I} ; %
  \draw [->] (c) to [out=40,in=140] (f);
  \draw [->] (c) to [out=40,in=140] (x_1);
  \draw [->] (c) to [out=40,in=140] (I);
  \draw [->] (x_0) to [out=40,in=140] (x_1);
  \draw [->] (x_0) to [out=40,in=140] (I);
  \draw [->] (f) to [out=40,in=140] (x_1);
  \draw [->] (f) to [out=40,in=140] (I);
\end{tikzpicture}}
\end{subfigure}
\end{minipage}
\caption{\textbf{Left:} The graphical model of the target distribution $p_{\theta}$ that our algorithm approximates. $c$ is a context and $x_1$ is a high-quality LM output and $\mathcal{I}$ indicates whether the output is high-quality according to human preferences. \textbf{Right:} The graphical model of the proposal distribution $q$ we use for importance sampling. $x_0$ is an initial LM output and $f$ is language feedback on $x_0$.} 
\label{fig:graphical_model_I}
\end{figure}
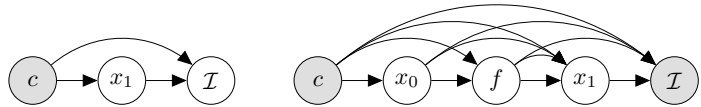

Our goal is to produce a high-quality output $x_1$ for a context $c \sim p(c)$ (e.g., a summary of a document). We use an LM $\pi_{\theta}$ to generate an output $x_1$, by conditioning on the context $c$, i.e., $x_1 \sim p_{\theta}(x_1|c)$. We then introduce the predicate $\mathcal{I}$, a random variable such that $\mathcal{I}=1$ if the output is high quality according to human preferences.
We denote this data-generating process, shown in Fig.~\ref{fig:graphical_model_I} left, as:
\begin{align}
    p_\theta(c, x_1, \mathcal{I}) = p(c) \pi_\theta(x_1|c) p(\mathcal{I}|c,x_1).
\end{align}
%The goal of maximizing summary helpfulness across contexts $c$ can then be framed as maximising the expected probability of helpfulness:
%\begin{align}
%p_\theta(\mathcal{I}=1) &= \mathbb{E}_{c \sim p(c)} p(\mathcal{I}=1|c) \\
%&= \mathbb{E}_{c \sim p(c)} \mathbb{E}_{x_1\sim \pi_\theta(x_1|c)} p(\mathcal{I}=1|c,x_1)
%\end{align}
%If we treat $p(\mathcal{I}=1|c, x_1)$ as a reward, this objective is equivalent to the RL objective of the expected reward. 
%However, instead of using RL, 
We frame our goal as maximizing the marginal log probability of quality across contexts: $\mathbb{E}_{c \sim p(c)} \log p(\mathcal{I}=1|c).$ %
% of generating high-quality summaries as 
% expected log probability of summary quality, i.e., $\mathbb{E}_{c \sim p(c)}  \mathbb{E}_{x_1\sim \pi_\theta(x_1|c)} \log p(\mathcal{I}=1|c,x_1)$.  
For a particular context $c$, we approximate $\log p(\mathcal{I}=1|c)$ by introducing an importance sampling proposal distribution $q(x_1|c)$ and using the Evidence Lower Bound (ELBo):
%We will use the fact that $p(\mathcal{I}=1|c)$ has the following lower bound: 
\begin{align}
\log p(\mathcal{I}=1|c) &= \log \sum_{x_1} p_\theta(x_1, \mathcal{I}=1|c) \label{eq:two} \\
% &= \log \sum_{x_1} q(x_1|c) \frac{p_\theta(x_1, \mathcal{I}=1|c)}{q(x_1|c)} \label{eq:three}\\
&\geq \sum_{x_1} q(x_1|c) \log \frac{p_\theta(x_1, \mathcal{I}=1|c)}{q(x_1|c)} \label{eq:four}
% &= \sum_{x_1} q(x_1|c)  \Big[ \log p_\theta(x_1, \mathcal{I}=1|c) - \log q(x_1|c) \Big] \label{eq:five} \\
%&:= F(\theta, q) \label{eq:six}
\end{align}
%In Eq.~\ref{eq:two}, we use the sum rule, in Eq.~\ref{eq:three}, we introduce an importance sampling proposal distribution $q(x_1|c)$, in Eq.~\ref{eq:four} we use the Jensen inequality and in Eq.~\ref{eq:six} we define $F$, the lower bound to maximize. 

We maximize the lower bound in Eq. \ref{eq:two}, henceforth called $F(\theta, q)$, using an Expectation-Maximization (EM) procedure: alternating between maximizing $F$ w.r.t. the proposal distribution $q$ (E-step) and w.r.t. $\pi_{\theta}$ (M-step)  We call this algorithm Imitation learning from Language Feedback.

 %
% There exists a closed form for such an optimal $q$ (see Appendix~\ref{appendix:optimal_q} for the full derivation):
% \begin{align*}
% \operatorname*{argmax}_q F(\theta, q)
% &= p_\theta(x_1|\mathcal{I}=1,c) \vcentcolon= q^*(x_1|c).
% \end{align*}
% \begin{figure}[t]
%     \centering
%     \begin{subfigure}
%         \input{images/dgp_q}
%         \caption{A graphical model of the data generating process underlying the proposal distribution $q^*$}
%         \label{fig:graphical_model}
%     \end{subfigure}
    % \begin{subfigure}
    %     \input{images/dgp_p}
    %     \caption{A graphical model of the data generating process underlying the proposal distribution $q^*$}
    %     \label{fig:graphical_model}
    % \end{subfigure}
% \end{figure}
% \paragraph{Approximating the Proposal Distribution}
% Even with a closed-form expression for $q^*$, we still do not have an easy way to sample from it. To sample from the posterior $p_{\theta} (x_1|\mathcal{I}=1,c)$, a distribution over high-quality texts, 
\paragraph{E-step}

Maximizing $F(\theta,q)$ w.r.t $q$ corresponds to refining the proposal distribution $q$ to assign higher likelihood to high-quality texts.
%We perform this refinement 
This is achieved by embedding $x_1$ into a data-generating process involving humans, by introducing the initial output $x_0$, and human feedback $f$ (via sum rule):
% and cast $\mathcal{I}$ in terms of human feedback:
\begin{align}
q(x_1|c) &= \sum_{x_0, f} p_{\theta}(x_0, f, x_1|\mathcal{I}=1,c) \label{eq:13}\\
&\propto\sum_{x_0, f}p_{\theta}(x_0,f,x_1|c)p_{\theta}(\mathcal{I}=1|c,x_0, f, x_1) \label{eq:14} \\
&= \sum_{x_0, f} p_{\theta}(x_0|c)p(f|c,x_0)p_{\theta}(x_1|c,x_0,f) \nonumber\\ & \quad\quad\quad p_{\theta}(\mathcal{I}=1|c,x_0, f, x_1).  \label{eq:15}  
\end{align}
%In Eq.~\ref{eq:13}, we introduce latent variables $x_0$, the initial summary, and $f$, the human feedback, using the sum rule. In Eq~\ref{eq:14}, we use Bayes' Rule, and in Eq.~\ref{eq:15}, we factorize the joint distribution $q(x_0,f,x_1)$ using the product rule
Eq. \ref{eq:15} gives rise to the following sampling procedure (see also Fig.~\ref{fig:graphical_model_I}, right): First, an LM is conditioned on the context $c$ and generates an initial output $x_0$. Second, a human provides language feedback $f$ on the $(c, x_0)$ pair. Third, the LM generates a refined text $x_1$ conditioned on $(c, x_0, f)$. Finally, a binary variable $\mathcal{I}$ indicates whether $x_1$ is a high-quality text, given an initial output $x_0$, feedback $f$, and a context $c$. We model $p_{\theta}(\mathcal{I}=1|c, x_0, f, x_1)$ as a Boltzmann distribution:
\begin{align}
    p_{\theta}(\mathcal{I}=1|c, x_0, f, x_1) \propto \exp(R(c,x_0,f,x_1)/\beta),\label{eq:boltzmann}
\end{align}
which uses a reward function $R$ defined in terms of four variables: $c, x_0,f,x_1$; $\beta$ is a temperature hyperparameter. This Boltzmann distribution makes quality easy to evaluate since it expresses it as a reward function $R$ of a previous output and human language feedback.

We now argue why the E-step results in a proposal distribution that is better than the original distribution $p_\theta(x_1|c)$, i.e., why samples from $q(x_1|c)$ tend to be of higher quality than samples from $p_\theta(x_1|c)$. First, we know that $x_0$ is already a reasonably good output (since $\pi_{\theta_\text{old}} \approx \pi_\theta$). We can assume that the feedback $f$ is informative and high-quality. Therefore $x_1 \sim p_\theta(x_1|c,x_0,f)$ is going to be of higher quality than $x_0 \sim p_\theta(x_0|c)$ because it leverages useful information from the feedback.  Furthermore, let us choose $R$ to assign higher values to refined texts $x_1$ that improve upon $x_0$ w.r.t to $f$ and $c$. Consequently, Eq.~\ref{eq:boltzmann} assigns a higher likelihood to high-quality outputs $x_1$, allowing us to put additional weight on high-quality outputs and improving the proposal distribution $q$ further.

% will result in proposal $q(x_1|c)$ being improved over $\pi_\theta(x_1|c)$. 

%\begin{enumerate}
   % \item A LM is conditioned on the document $c$ and generates initial summary $x_0$,
  %  \item A human provides language feedback $f$ on the $(c, x_0)$ pair,
   % \item A LM generates a refined summary $x_1$ conditioned on $(c, x_0, f)$,
   % \item Finally, a binary variable $\mathcal{I}$ indicates whether $x_1$ improves upon $x_0$ with respect to feedback $f$ and source document $c$.
%\end{enumerate}

% This data-generating process gives rise to the following joint distribution:
% \begin{align}
%     q(x_0, f, x_1, \mathcal{I}|c) = &q(x_0|c) q(f|c, x_0)  \\ &q(x_1|f, x_0, c) q(\mathcal{I}|c, x_0, f, x_1) \nonumber
% \end{align}

\paragraph{M-step}

Maximizing $F(\theta,q)$ w.r.t. the policy $\pi_{\theta}$ is equivalent to supervised learning (minimizing cross-entropy loss) on a distribution defined by $q$. To see that, we drop all the terms from Eq.~\ref{eq:four} that do not depend on $\theta$:
\begin{align}
\operatorname*{argmax}_\theta F(\theta, q) 
&= \operatorname*{argmax}_\theta \mathbb{E}_{x_1 \sim q(x_1|c)}  \log p_\theta(x_1,\mathcal{I}=1|c) \nonumber \\
 &= \operatorname*{argmin}_\theta \mathbb{E}_{x_1 \sim q(x_1|c)} -\log \pi_\theta(x_1|c ).\label{eq:supervised_finetuning} \hspace{-10px}
\end{align}
 %In Eq.~\ref{eq:eight} we decompose $\log p_{\theta}(x_1,I=1|c)$, drop $\log p(\mathcal{I}=1|c, x_1)$ as it does not depend on $\theta$ and switch from maximization to minimization.

\paragraph{ILF: Imitation learning from Language Feedback}
\label{sec:ilf_reward}

In ILF, we alternate between the E-step and M-step, using the pseudocode in Algorithm~\ref{alg:expert_iteration}. In the M-step, we use the model from the previous iteration $\pi_{\theta_\text{old}}$ as both $p_{\theta}(x_0|c)$ and $p_{\theta}(x_1|c,x_0,f)$. In practice, we implement $R$ by conditioning an instruction-finetuned LM on a binary question such as \textit{Does this new text incorporate the feedback provided? Answer Yes or No.} where the label $y$ is either $y_{\text{good}}$ (`` Yes") or $y_{\text{bad}}$ (`` No"). We use the probability of the positive answer $y_{\text{good}}$ given the prompt as a reward, i.e. $p(y_{\text{good}}|\text{prompt}) = \frac{p(y_{\text{good}}|\text{prompt})}{p(y_{\text{good}}|\text{prompt}) + p(y_{\text{bad}}|\text{prompt})}$. With these assumptions, $q$ takes the form:
\begin{align*}
   q(x_1|c) \propto ~ &\mathbb{E}_{x_0 \sim \pi_{\theta_\text{old}}(x_0|c)} \mathbb{E}_{f\sim p(f|c,x_0)} \\ &\pi_{\theta_\text{old}}(x_1|c,x_0,f) 
   \exp(R(c,x_0,f,x_1)/\beta).\nonumber
\end{align*}
We take advantage of this proposal distribution and perform the M-step, i.e., $\text{argmax}_{\theta} F(\theta, q)$ on optimized data. 
Finally, we approximate sampling from $q(x_1|c)$ by best-of-$N$ sampling. To obtain a sample $x_1 \sim q$, we %
% first sample $x_0 \sim \pi_{\theta_\text{old}}(x_0|c)$ and $f \sim p(f|c,x_0)$, then
sample $N$ refinements $\{x_1^1, \dots, x_1^N \} \sim \pi_{\theta_\text{old}}(x_1|c,x_0,f)$, and compute
\begin{align*}
x_1 = \text{argmax}_{x_1^i} \exp R(c,x_0, f, x_1^i).
\end{align*}

In summary, we show that ILF can be understood as Bayesian inference. This process involves updating an LM based on the evidence provided by language feedback. This lens highlights the correspondence between ILF and RL with Human Feedback ~\citep[][\textit{inter alia}]{ziegler2019fine, stiennon2020learning}, which was previously demonstrated to be equivalent to Bayesian inference~\citep{korbak2022rl}. 

%\subsection{Optimal proposal distribution}\label{appendix:optimal_q}

%\begin{align}
%\text{argmax}_q F(\theta, q)
%&= \text{argmin}_q \sum_{x_1|c} q(x_1|c) \log \frac{q(x_1|c)}{p_\theta(x_1,\mathcal{I}=1|c)} \label{eq:nine}\\
%&= \text{argmin}_q \sum_{x_1|c} q(x_1|c) \log \frac{q(x_1|c)}{p_\theta(x_1|c,\mathcal{I}=1)} \label{eq:ten}\\
%&= \text{argmin}_q D_\text{KL} \Big( q(x|c), p_\theta(x_1|c,\mathcal{I}=1) \Big) \label{eq:eleven}\\
%&= p_\theta(x_1|c, \mathcal{I}=1) .\label{eq:twelve}
%\end{align}

%In Eq.~\ref{eq:nine}, we add a minus in front of Eq.~\ref{eq:four}, which flips the fraction in the logarithm and switches from maximization to minimization. In Eq.~\ref{eq:ten}, we multiply by the constant $\log p(\mathcal{I}=1|c)$, which converts the joint distribution in the denominator to a conditional distribution. In Eq.~\ref{eq:eleven}, we use the definition of the Kullback-Leibler (KL) divergence, and in Eq.~\ref{eq:twelve} use the fact that $D_{KL}(a,b)$ is minimized at $a=b$. Therefore, we arrive at a closed-form solution for the optimal proposal distribution, which we call $q*$.

% \begin{figure}[h!]
%     \begin{subfigure}{0.5\textwidth}
%     \centering
%            \input{images/dgp_p}
%          \label{fig:dgp_p}
% \end{subfigure}%
% \begin{subfigure}{0.5\textwidth}
% \centering
%         \input{images/dgp_q}
%          \label{fig:dgp_q}  
% \end{subfigure}
%  \caption{\textbf{Top:} Target distribution $p_{\theta}$. \textbf{Bottom:} Proposal distribution $q^*$.}
% \label{fig:data_generating_process}
% \end{figure}

\section{Additional Related Work on Language in RL Settings}
\label{app:releated_works_rl}
Language has been widely used in RL for various purposes~\citep[see][for an overview]{luketina2019survey}, such as specifying tasks~\citep[``instruction following''][\textit{inter alia}], driving exploration~\citep{tam2022semantic}, inferring reward functions~\citep[][\textit{inter alia}]{lin2022inferring, sumers2021learning, fidler2017teaching}, and training a model via strong supervision~\cite{andreas2017modular,kaplan2017beating}, reward shaping~\cite{goyal2019using}, or by providing descriptions of trajectories \citep{nguyen2021interactive}. In contrast, we use language to correct faulty behavior. Other work uses language feedback at test time to correct mistakes in a model's behavior, e.g., image segmentation~\citep{rupprecht2018guide} or code generation~\cite{elgohary-etal-2020-speak,austin2021program}.
In contrast, we use feedback to \textit{train} models, and our approach does not require human intervention at test time.

\section{Dataset Collection and Analysis}
\label{app:dataset}

\paragraph{Annotation process}
To ensure the high quality of our human annotations, we employ experienced annotators sourced through the data-labeling company Surge AI. During an onboarding and evaluation process, we calculate author-annotator agreement on the binary comparison task and manually review the quality of the written feedback and ideal summaries to ensure their high quality. 
Then we select 31 qualified annotators for all annotation tasks, though they can choose which tasks to participate in and for how long. To further ensure the quality of our annotations, we provide detailed instructions, which we provide to the annotators, and update throughout the process to ensure continuous improvement (these instructions can be found in Appendix~\ref{app:annotator_instructions}). To measure the agreement rate between the annotators and the authors, we select a sample of 10 Reddit posts from the training dataset as a gold standard and have 17 annotators label them. When comparing the binary comparison annotations with our own ones, this results in an author-annotator agreement rate of $81.0 \%$. We also calculate the average agreement rate between all the possible annotator combinations, yielding an annotator-annotator agreement of $70\%$. By utilizing these thorough processes and evaluations, we can ensure the accuracy and reliability of our human annotations.

%We hire annotators through the data-labeling startup Surge AI\footnote{\url{https://surgehq.ai}}. We first onboard the annotators and evaluate their annotation quality, after which we pick 31 annotators (a few annotators were later removed, due to poor annotation quality). For all annotation tasks, only those 31 annotators were allowed to work on them, however, they were free to choose which tasks they wanted to participate in and for how long. We evaluate the annotation quality by measuring the agreement rate with the authors on the binary comparison task, and by manually reviewing the quality of the written feedback and human-written, ideal summaries. We use detailed procedures and instructions to ensure high agreement between the annotators and us, on the given task (See Appendix~\ref{app:annotator_instructions} for the instructions we provided to the annotators). We provide feedback to the annotators and update instructions throughout the process to ensure continuous improvement in the annotation quality. For our largest annotation, the training dataset, we selected $10$ samples as the gold standard, on which we indicated our binary preferences. We then had $17$ annotators label those samples, which results in an author-annotator agreement of 81$\%$.

\paragraph{Dataset Analysis}
\label{sec:dataset_analysis}
The feedback we collect typically addresses the most critical shortcomings of the summaries. In $92.0 \%$ of our train samples, the annotators' feedback was complete and addressed all important shortcomings of the summary, as reported by the annotators. Across our train dataset, we observe that the majority of the feedback pertains to coverage ($77.0 \%$), with smaller percentages relating to accuracy ($16.0 \%$), coherence ($5.0 \%$), and other categories ($2.0 \%$). We also analyze the length of the various summaries and feedback, measured in the average number of tokens. Our human-written summaries have an average length of $41.0 \pm 0.1$ tokens, the extracted human summaries from Reddit had an average length of $32.5 \pm 0.1$ tokens, the initial summaries generated by FeedME have an average length of $29.3 \pm 0.1$ tokens, and the feedback written by annotators on these initial summaries has an average length of $20.4 \pm 0.2$ tokens.

\begin{figure}[bh!]
\centering
%\begin{subfigure}{.3\textwidth}
    \includegraphics[valign=t, scale=0.4]{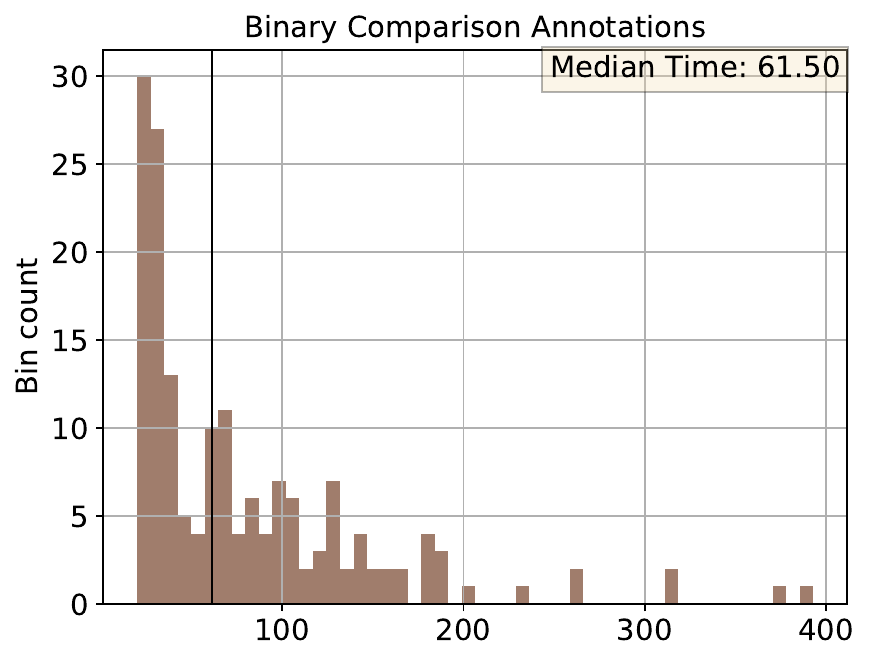}
  %  \label{fig:annotation_time_comparison}
%\end{subfigure}
\hfill
%\begin{subfigure}{.3\textwidth}
    \includegraphics[valign=t,scale=0.4]{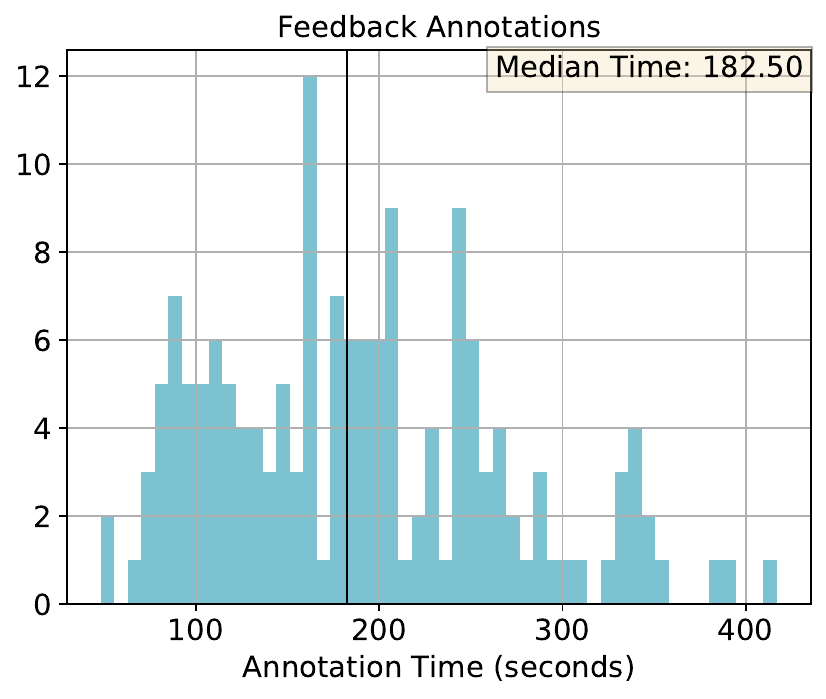}
   % \label{fig:annotatino_time_feedback}
%\end{subfigure}
\hfill
%\begin{subfigure}{.3\textwidth}
    \includegraphics[valign=t, scale=0.4]{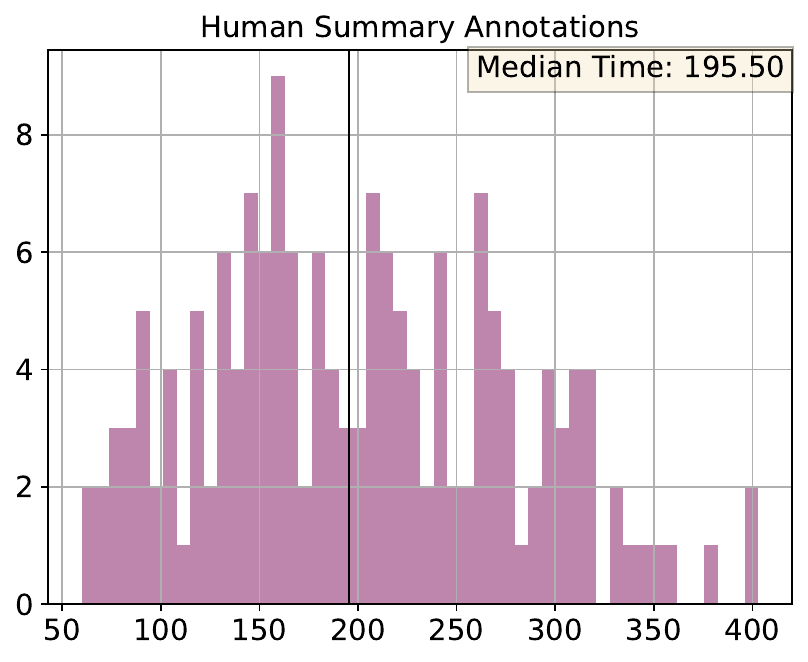}
  %  \label{fig:annotation_time_ideal}
%\end{subfigure}
\caption{Histogram Plot of annotation times (in seconds) of the binary comparison task, the feedback annotation task and the human summary writing task. The evaluation is conducted on the development dataset. We observe that annotators are much quicker at the binary comparison task, which is expected. The results also show that writing feedback takes less time than writing an ideal summary.}
\label{fig:annotation_times}
\end{figure}

In addition to these analyses, we also measure the time it takes annotators to complete various tasks (i.e., binary comparison, feedback writing, and ideal summary writing) on our development dataset. We ignore outliers and consider only samples with annotation times of at least 20 seconds and at most 420 seconds (7 minutes). Annotators take $61.5 \pm 5.3$ seconds on average on the binary comparison task, $182.5\pm 6.3$ seconds on the feedback task, and $195.5\pm 6.1$ seconds on the ideal summary task.
 We plot the annotation times on the development dataset for the tasks of annotating binary comparisons, writing feedback, and writing ideal summaries as histograms in Fig.~\ref{fig:annotation_times}. The annotators are much faster at annotating binary comparisons than feedback or ideal summaries. Writing feedback takes less time than writing ideal summaries, which is expected, as critiquing a task is usually easier than solving it. These comprehensive evaluations demonstrate the high quality and thoroughness of our dataset and annotation processes.

\section{Targeted Word Removal Details}
\label{sec:word_removal_example}
Below is an example of how we instruct or ``prompt'' an LM to remove specific, offensive words from a sentence.
\begin{quote}
    \textit{``In this text, many toxic and offensive words are used: You are such a jerk, and a nice person, and an idiot. The ideal text should remove the word jerk, but otherwise be unchanged: You are''}
\end{quote}
Here, the target completion is \textit{`` such a nice person and an idiot.''}
More formally, we sample offensive sentences by using $k$ offensive words from a fixed set of 25 offensive words drawn uniformly at random (without replacement). Each offensive sentence also includes the words "nice person" in addition to all the offensive words. For each $k \in \{1, \dots, 10\}$, we sample 50 offensive sentences. The task is then to remove $l \in [1, 2, 3]$ offensive words from a given sentence with $k \geq l$. Since we include the words "nice person" in the offensive sentence, we can remove $l=k$ offensive words and still have a target sentence that intuitively makes sense.

\section{Details about Ranking Procedure}
\label{app:ranking}
We use a standard ranking scheme where each of $K$ summaries is given a rank between 1 and $K$ (inclusive). Sometimes refinements are exact copies of the initial summaries or are very similar in terms of quality, which is why we allow for summaries to be tied. When calclating the win rate we assign $0.5$ wins for tied samples. We assign the rank $r'$ to all summaries ranked in a tie, where $r'=\frac{r + (r+n-1)}{2}$, $r$ is the rank of the tied elements, and $n$ is the number of ties at the rank. For example, we map a ranking of $(1,2,2,4,5) \rightarrow (1,2.5,2.5,4,5)$ and a ranking of $(1,2,3,3,3) \rightarrow (1,2,4,4,4)$.

\section{Reward Model}
\label{app:reward_model}
Here we describe the various RMs that we evaluate in more detail. We evaluate the final RM that we use, which produces a language output (e.g., `` Yes" or `` No") and a standard reward model that produces a scalar output.

\paragraph{Standard RM.}
Akin to \cite{stiennon2020learning}, we remove the last embedding layer of a language model and train it to output a scalar value. This scalar value predicts which summary, $x \in {\{x_0^0, x_0^1\}}$, is better as judged by a human, given a context $c$. We use the OPT 13B LM, introduced in \citep{zhang2022opt}, as the base model for our RM and finetune it on the human preference comparisons that we collected. It is worth noting that it is not possible to add linear layers on top of GPT-3 models provided via the API, which is why we use the OPT model.

%If the summary preferred by the human is $x_0^i$, the standard RM loss can be defined as:
%$$\text{loss}_{\text{base}}(r_{\theta})= - \mathbb{E}_{(c, x_0^0, x_0^1, i) \sim D} [\log (\sigma (r_{\theta}(c, x_0^i) - r_{\theta}(c, x_0^{1-i})))] $$

\paragraph{Reward Model with Language Output.}
In addition to the classic RM~\citep{stiennon2020learning},
we train an RM to output language tokens instead of a scalar value. To do so, we finetune an LM to classify whether a summary $x_0$ is high quality or not, by training it to predict a label $y \in \{ y_{good}, y_{bad} \}$.
We then finetune the LM to maximize $\lambda \log p(x_0) + \log p(y | x_0)$, where $\lambda \in [0,1]$, chosen using the development dataset. The complete loss can also be written as: 
$$\mathcal{L}(p_{\theta}, x, y) = - \lambda \cdot \sum_{t=1}^{|x|} \log p_{\theta} (x_t|x_{<t}) - \sum_{t=1}^{|y|}\log p_{\theta} (y_t|x, y_{<t}).$$
where the subscript $t$ indicates the token index.
We evaluate the finetuned LM on a given summary $x_0$ by computing $p(y_{good}|x_0)$. The best RM overall uses the following instruction \textit{Is the
above an excellent summary of the given text? An excellent summary is coherent, accurate, concise, and detailed. Answer with Yes or No.}, which we refer to as the OPT-RM (when finetuning OPT-13B) and GPT-3 Binary (when finetuning GPT-3-175B). We also explore finetuning on another prompt, where we provide both summaries $A$ and $B$ to the LM and instruct it to indicate which summary is preferred, i.e. \textit{Question: Which summary is the better one? An excellent summary is coherent, accurate, concise, and detailed. Answer with A or B.} We then finetune the LM on the label of the preferred summary (according to binary human feedback), i.e. on $y \in \{ y_{A}, y_{B} \}$. We evaluate the finetuned LM on a given summary $x_0$ by computing $p(y_{A}|x_0)$. We refer to this RM as \textit{Comparison} RM.  We explore two RMs, namely, OPT-13B \citet{zhang2022opt}, and GPT-3-175B and refer to Appendix~\ref{app:hp_tuning} for the hyperparameters we use and to Appendix~\ref{app:rm_prompts} for the prompt templates).

\paragraph{Results.}
We evaluate all RMs on our validation dataset, and calculate the accuracy of predicting the preferred summary out of two, based on human preferences. Table~\ref{tab:reward_model_accuracy_development} shows the complete results, and here we report on some of the RMs trained on 5K samples. The OPT model with the standard RM loss achieves an accuracy of $71.8 \pm 2.0 \%$ on the validation dataset. The results further show that both of our methods for training OPT with the LM loss outperform the standard RM loss, with OPT comparison achieving an accuracy of $72.6 \pm 1.9 \%$, and OPT-RM an accuracy of $73.4 \pm 1.9 \%$. We obtain similar results with finetuning GPT-3-175B, achieving an accuracy of $71.2 \pm 2.0 \%$ with the GPT3 Comparison, and an accuracy of $74.2 \pm 2.0 \%$ with GPT-3 Binary, which outperforms the OPT-RM. %It is worth noting that we have more fine-grained control over various hyperparameters when training OPT compared to using OpenAI's API, making it difficult to directly compare the performance of both models and draw concrete conclusions.

\begin{figure}[t!]
\begin{center}
 \small{
   \cblock{191}{134}{173} OPT-RM, LM Loss/Binary\quad
   }
\end{center}
    \centering \includegraphics[width=.48\textwidth,keepaspectratio]{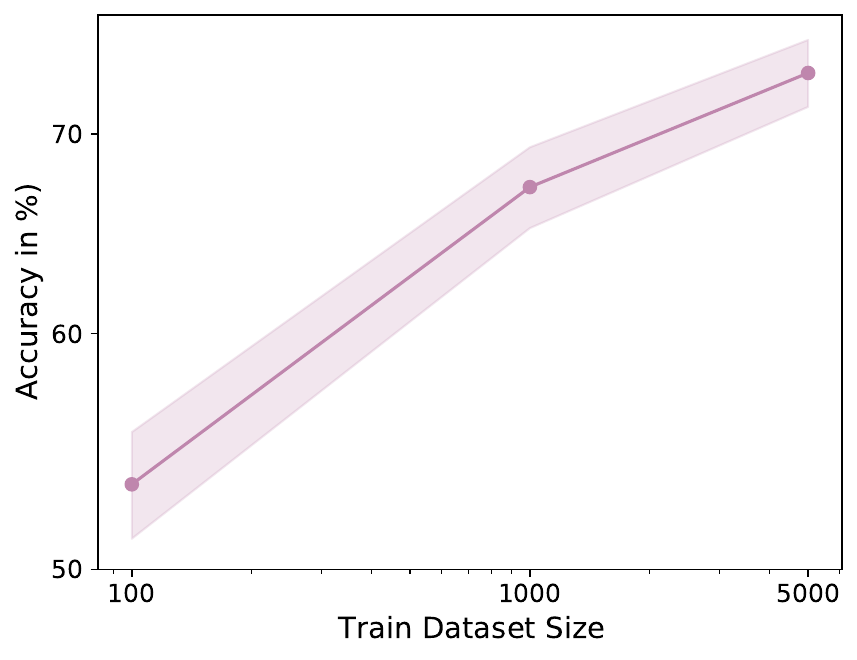}
    \caption{Here we plot the validation accuracy of OPT-RM trained on 100, 1K, and 5K samples on a log-log plot. The figure shows scaling when increasing the dataset size.}
    \label{fig:opt_reward_model_scaling}
\end{figure}

Based on these results, we further evaluate the OPT Binary and GPT-3-175B Binary models on the development dataset that we use to evaluate the scoring functions in \S\ref{sec:scoring_functions}. We calculate the fraction of times the refinement selected by an RM is better than a randomly-selected refinement (``win rate"), according to a ranking given by human evaluators (see App.~\ref{app:ranking} for more details). The results can be found in Table~\ref{tab:scoring_function_results_with_opt}. OPT-RM achieves a win rate of $63.3 \pm 2.7 \%$, and the GPT-3-175B Binary model achieved a win rate of $61.8 \pm 2.9 \%$. In this evaluation, OPT-RM outperforms GPT-3 Binary. When considering the results from both the validation and development datasets, both OPT-RM and GPT-3-Binary seem to perform similarly. Given that we have more control over the training process of OPT, the possibility of releasing the model, and the cost involved in training using OpenAI's API, we select OPT-RM model as our reward model for comparison with ILF.
In Figure~\ref{fig:opt_reward_model_scaling}, we show the validation accuracy of OPT-RM trained on 100, 1K, and 5K samples on a log-log plot. The figure shows scaling when increasing the dataset size.

 We further evaluate results for finetuning OPT-RM on the dataset of \citet{stiennon2020learning}, and also evaluating their model with 1.3B parameters on our dataset. We observe that the binary preference distribution of the training dataset has a significant impact on the performance of the reward model. For example, OPT-RM trained on 5K samples of our own train dataset (i.e., our final reward model) achieves an accuracy of $61.9 \pm 0.2 \%$ on the test set from \citet{stiennon2020learning} (not shown in Table~\ref{tab:reward_model_accuracy_development}). When this same model is trained on 90K samples from the train dataset of \citet{stiennon2020learning}, it achieves an accuracy of $69.3 \pm 0.2 \%$ on their test set (also not shown in Table~\ref{tab:reward_model_accuracy_development}). In contrast, this same model trained on 90K samples from their train dataset achieves an accuracy of only $68.6 \pm 2.0 \%$ on our validation dataset, which is significantly lower than the accuracy of $73.4 \pm 1.9 \%$ achieved by the model trained on 5K samples of our own train dataset. Similar patterns can be observed when comparing the OPT Binary model with 1.3B parameters trained on 5K samples of our own train dataset to the released 1.3B reward model trained by \citet{stiennon2020learning} on approx. 64K samples of their own train dataset. The former model achieves an accuracy of $69.6 \pm 2.0 \%$ on our validation dataset, while the latter only achieves an accuracy of $63.8 \pm 2.1 \%$ (note, though, that the RMs are trained with different loss functions).
These results highlight two important considerations: (1) preference distributions can vary significantly and have a strong effect on what a reward model learns, and (2) the sample efficiency of a reward model depends heavily on the train and test distributions. If the test distribution differs from the train distribution, reward models may be very sample inefficient and fail to accurately learn the true distribution, even when given significantly more samples.

\begin{table*}[t!]
\centering
\begin{tabular}{c c c} \toprule
 & Scoring Function &  \makecell{Win Rate vs Random Selection (in \%)} \\
\toprule
Task Specific Heuristic & Max Length & $65.0 \pm  2.7$\\
\midrule
\midrule
\multirow{2}{*}{Zero-Shot} & 
Embedding Similarity   &  
$48.3 \pm 3.0$\\
& \textbf{InstructRM Ensemble} & \textbf{56.0} $\pm$ \textbf{3.0}\\
\midrule
\multirow{2}{*}{Finetuning on 5K samples} & \textbf{OPT Binary} &  \textbf{63.3} $\pm$ \textbf{2.7}\\
& GPT-3 Binary & $61.8 \pm 2.9$\\
\bottomrule
\end{tabular}%}
\caption{In a human evaluation, we compare reward models and ranking methods on the development dataset (in the same way as in Fig~\ref{tab:scoring_function_results}. Both RMs are trained on 5K samples and outperform the zero-shot methods.}
\label{tab:scoring_function_results_with_opt} 
\end{table*}

\begin{table*}[t!]\resizebox{\textwidth}{!}
{
\begin{tabular}{c c c c c c c} \toprule
& Models  & \makecell{\# Params} & \makecell{Train Data Size} & \makecell{Development Accuracy (in \%)} & \makecell{Validation Accuracy (in \%)} \\
\midrule
\multirow{7}{*}{LM Loss / Our dataset}
& OPT Comparison & 13B& 5K & $66.5 \pm 3.3$&$72.6 \pm 1.9$ &\\
& OPT RM  & 1.3B & 5K & $70.0 \pm 3.2 $&$69.6 $$\pm 2.0$ &\\
& OPT RM  & 13B & 100 & $54.5 \pm 3.5$&$53.4 \pm 2.2$&\\
& OPT RM  & 13B & 1K & $68.5 \pm 3.2$& $67.2 \pm 2.1$&\\
& \textbf{OPT RM} & \textbf{13B} &\textbf{5K} & \textbf{69.5} $\pm$ 
\textbf{3.2}& \textbf{73.4} $\pm$ \textbf{1.9} &\\
& GPT-3 Comparison  & - & 5K&68.0  & $71.2 \pm 2.0$ &\\
& \textbf{GPT-3 Binary}  & - & \textbf{5K} & - & \textbf{74.2} $\pm$ \textbf{2.0} &\\
\midrule
RM Loss / Our dataset & 
OPT  & 13B & 5K &$68.5 \pm 3.2$ &$71.8 \pm 2.0$ &\\
RM Loss / \citet{stiennon2020learning} train dataset  & \citet{stiennon2020learning} RM & 1.3B & 64K & $58.0 \pm 3.4$ & $63.8 \pm 2.1$ &\\
\midrule
LM Loss / \citet{stiennon2020learning} train dataset & OPT Binary & 13B & 90K & $69.0 \pm 3.2 $&$68.6 \pm 2.0$&\\

\hline
 \bottomrule
\end{tabular}}
\caption{In a human evaluation, we evaluate various RMs on the development dataset and validation dataset. We also report the results of training on the train dataset of \citet{stiennon2020learning} and evaluating on our development and validation datasets. We calculate the accuracy of predicting which of two summaries is preferred by a human.}
\label{tab:reward_model_accuracy_development}
\end{table*}

\section{Hyper Parameters}
\label{app:hp_tuning}

\subsection{Generating Refinements}
\label{app:postprocessing}
For the targeted word removal experiments (\S\ref{sec:targeted_word_removal}), we use greedy decoding until 200 tokens or \textit{/\ n} is generated. For all summarization experiments we sample up to 48 tokens~\citep[as in][]{stiennon2020learning} with nucleus sampling~\cite{holtzman2019curious} with $p=0.95$ and temperature $t=1.0$. We strip non-alphanumeric characters (e.g., newlines) from the beginning of sampled summaries. We further remove empty white spaces in the generated summaries and remove all text that comes after a new line token \textit{/\ n}. Due to the maximum token length, sampled summaries sometimes end with incomplete sentences. Thus, we remove ending sentences that do not end in ``.'', ``!'', or ``?''. The described temperature and post-processing are applied to all summary generations, i.e., for generating initial summaries, refinements, and test summaries. 

\subsection{Finetuning on Summaries}
We conduct independent hyperparameter optimization sweeps with three dataset sizes of human summaries of 100, 1K and 5K samples, and then use the same hyperparameters for finetuning on refinements (ILF) and finetuning on initial summaries. We choose to run the hyperparameter sweep on Human summaries since this will not give an unfair advantage to our algorithm that finetunes on refinements. For the sweep, we utilize the train dataset of human summaries (consisting of 100, 1K, and 5K samples) and evaluate on the development dataset. Unfortunately, the OpenAI API only provides validation loss and token accuracy for batches of the development dataset, making it impossible to evaluate the model on the full development dataset during training. As a result, we utilize the model API to evaluate on the full development dataset after finetuning and calculate the perplexity of the generated summaries as a performance measure.

To determine the optimal hyperparameters, we perform a sweep over a range of values for the following parameters: \textit{epochs} $\{1,2,3,4\}$, \textit{prompt loss weight} $\{0,0.01,0.05,0.1\}$, and \textit{learning rates} $\{0.02,0.05,0.1,0.2\}$. We first sweep over epochs and select the best value, then perform a sweep using that value for the prompt loss weight, and so on. Our empirical observations indicate that the number of epochs has the greatest impact on perplexity, with training for more than one epoch resulting in overfitting. The selected hyperparameters can be found in Table~\ref{tab:finetuning_hp}.

During the finetuning phase for the \textsc{Refinements} and \textsc{Initial Summaries} datasets with 1K samples each, we made an error in our hyperparameter selection. Instead of using a prompt loss weight of $0.05$, we mistakenly used a value of 0, when finetuning on human summaries. While this error may have slightly impacted our results, the difference in perplexity between the two settings is minimal, with a value of $6.68$ for a prompt loss weight of $0.05$ and $6.71$ for a prompt loss weight of 0. Despite this mistake, our method still outperforms finetuning on human summaries for 1K samples, as well as finetuning on initial summaries using suboptimal hyperparameters.

\begin{table}[t!]
\centering
{
\begin{tabular}{c c c c c} \toprule
Samples  & \makecell{Epochs} & \makecell{Prompt Loss Weight} & \makecell{Learning Rate} \\
\hline
100 & $1$ & $0$ &  $0.05$ \\
1K & $1$ & $0.05*$ &  $0.02$ \\
5K & $1$ & $0.1$ &  $0.2$ \\

 \bottomrule
\end{tabular}}
\caption{We report the chosen hyperparameters of finetuning on 100, 1K, and 5K samples of \textsc{Human Summaries}.\newline
*This hyperparameter is optimal but used only for finetuning on \textsc{Human Summaries}. For finetuning on \textsc{Refinements} and \textsc{Initial Summaries} we inadvertently use the prompt loss weight 0.}
\label{tab:finetuning_hp}
\end{table}

\subsection{Multiple Iterations of ILF}
To evaluate multiple iterations of ILF, i.e., multiple iterations of refining-and-finetuning, we finetune GPT-3-175B on a refinement dataset with 200 and 300 samples. Thus we conduct a hyperparameter optimization on a train dataset of 200 and 300 refinements and evaluate on a development dataset of 200 refinements (instead of human summaries). To determine the optimal hyperparameters, we perform a sweep over a range of values for the following parameters: \textit{epochs} $\{1,2,3,4\}$, \textit{prompt loss weight} $\{0,0.01,0.05,0.1\}$, and \textit{learning rates} $\{0.02,0.05,0.1,0.2\}$. We first sweep over epochs and select the best value, then perform a sweep using that value for the prompt loss weight, and so on. For finetuning on 200 refinements we select the following hyperparameters: $\text{epochs}=1$, $\text{prompt loss weight}=0.05$, $\text{learning rate multiplier}= 0.1$. For finetuning on 300 refinements we select $\text{epochs}=1$, $\text{prompt loss weight}=0$, and $\text{learning rate multiplier}=0.2$.

\subsection{Finetuning Reward Models}

\paragraph{OPT Reward Model.}
% jon can you please fill this out.
For finetuning the OPT Reward Model, we perform bayesian hyperparameter optimization for each of the three different types of reward models:  \textit{Standard}, \textit{Comparison} and \textit{Classification} (see section \ref{app:reward_model}). We sweep over the learning rate in the range of $[1\text{e}^{-5}, 1\text{e}^{-6}]$ and the batch size $\{32, 64\}$ for all the models. For the reward models using the language loss, we also optimize the prompt-loss weight $\{0.0, 0.01, 0.05, 0.1, 0.5, 1.0\}$. We run 10 iterations per model and evaluate all the sweeps with the 200 development examples. We use a linear learning rate scheduler and a weight decay of $0.1$ for all the runs. The optimal batch size is 32 for all the models. The best prompt loss weight is $0.01$ for both the \textit{Comparison} and \textit{Classification} RMs. As for the learning rate,  we use $9.3\text{e}^{-6}$ for the \textit{Standard} RM, $5.8\text{e}^{-6}$ for the \textit{Classification} RM and $1\text{e}^{-6}$ for the \textit{Comparison} RM. In the final finetuning, we select the best RM in the validation split over 10 epochs. 

\paragraph{GPT-3 Reward Model.}
In order to finetune GPT-3-175B as an RM, we utilize the OpenAI API. We finetune two types of RMs: the \textit{Comparison} RM, which learns to predict which of two summaries is superior, and the \textit{Classification} RM, which predicts whether a given summary is of high quality or not. For cost considerations, we conduct hyperparameter tuning on a training dataset of 1K samples (instead of 5K) and evaluate on a development dataset of 200 samples. We use a dataset with 1K samples for cost reasons. We then apply the same hyperparameters when finetuning on 5K samples while implementing early stopping in terms of epochs. Due to the binary nature of the human preference annotations in the classification reward model, the effective train dataset size for this model is doubled to 2K samples.

In order to determine the optimal hyperparameters, we perform a sweep over a range of values for the number of epochs $\{1,2,3,4\}$ and the prompt loss weights $\{0, 0.001, 0.005, 0.01, 0.05, 0.1, 0.5\}$. The OpenAI API provides classification accuracy (for both the comparison and classification tasks) for the full development dataset after each epoch, allowing us to select the appropriate number of epochs and prompt loss weight. When finetuning on 5K samples, we utilize early stopping to prevent overfitting, using 1 epoch and a prompt loss weight of 0 for the comparison model and 4 epochs and a prompt loss weight of $0.001$ for the classification model. We use default values for all other hyperparameters, which may vary depending on the dataset size.

\begin{figure}[t!]
\begin{center}
     \small{
   \cblock{125}{194}{209} Finetuned on Human Summaries\quad
   \cblock{206}{119}{107} Finetuned on Refinements\quad
   \cblock{243}{192}{125} Finetuned on Initial Summaries\quad
   }
   \end{center}

    \centering \includegraphics[scale=0.6]{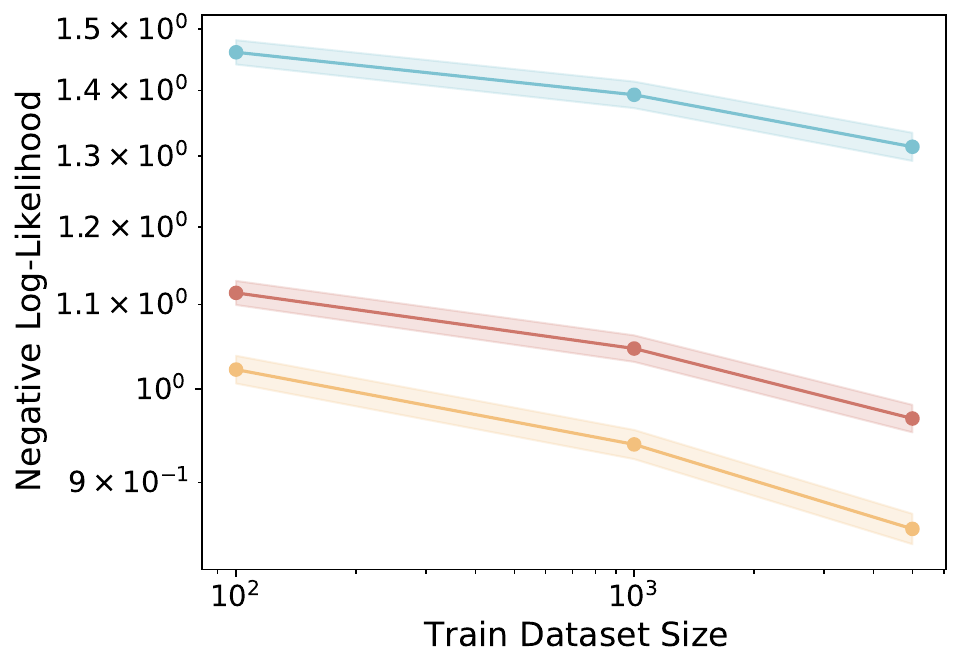}
    \caption{Evaluation of models finetuned on 5K initial summaries, refinements, and human summaries on 500 samples from the corresponding validation datasets. For example, the model finetuned on human summaries is evaluated on 500 human summaries from the validation dataset. The model finetuned on refinements has a significantly lower negative log-likelihood than the model finetuned on human summaries.}
    \label{fig:finetuned_methods_loss_scaling}
\end{figure}

\section{Additional Results}
\subsection{Analyis of Finetuned Models}
In Table~\ref{tab:Kl_distance_of_finetuned_models}, we evaluate GPT-3-175B on various finetuning datasets used for finetuning: the refinements, the initial summaries, and the human summaries. We evaluate the log-likelihood of GPT-3-175B on the summaries of 1K samples from the various train datasets (i.e. initial summaries, refinements, and human summaries). Concretely, we pass the whole prompt to GPT-3-175B, including the Reddit post, but only evaluate the log-likelihood of the completion, i.e. the generated summary. We also measure the (reverse) KL divergence \citep[following][]{gao2022scaling}  between an ILF-finetuned model and the pretrained LM before ILF-training, $D_{\text{KL}}(\text{finetuned}|\text{GPT-3-175B})$. We sample unconditionally (i.e. using a beginning of sentence token) from the finetuned models and evaluate the log-likelihood of the generated text with GPT-3-175B. We also report the forward KL divergence, $D_{\text{KL}}(\text{GPT-3-175B}| \text{finetuned})$. We discuss the results in \S \ref{sec:results_distribution}.

\begin{table*}[t!]\resizebox{\textwidth}{!}{
\begin{tabular}{c c c c} \toprule
Model &  \makecell{Neg. Log Likelihood of GPT-3-175B \\ on 1K train samples of respective distribution} &\makecell{$D_{KL}(\text{GPT-3-175B} | \text{finetuned})$ (in nats)} & \makecell{$D_{KL}(\text{finetuned}|\text{GPT-3-175B} )$ (in nats)}\\
\hline
Finetuned on Initial Summaries  &  $1.19 \pm 0.01 $&$0.43 \pm 0.11$ & $0.83 \pm 0.08$  \\
Finetuned on Refinements &$1.37 \pm 0.01 $&$0.60 \pm 0.10 $& $1.10 \pm 0.06$ \\
Finetuned on Human Summaries & $1.61 \pm 0.01$ &$0.12 \pm 0.09$ &$0.55 \pm 0.01$ \\
OPT-RM best-of-64 FeedME & - &- & $3.17$ \\
\hline
\bottomrule
\end{tabular}}
\caption{First we evaluate the log-likelihood of GPT-3-175B on the 1K samples of the various data distributions that we finetune on. Then we empirically calculate the KL-divergence by sampling 2000 texts of length 64 tokens from GPT-3-175B and evaluating the log-likelihood of the finetuned models on the samples (for the reverse KL we sample from the finetuned models and evaluate GPT-3-175B on the samples). We report the mean and standard error across 2 runs. For Best of 64 on a specific reward model, we use the analytical formula $KL(N, RM)= \log N - \frac{N-1}{N}$ (see also \cite{goodhart_gao}).}
\label{tab:Kl_distance_of_finetuned_models} 
\end{table*}

\subsection{Results: ILF + OPT-RM}
In this section, we present the full results of our best-performing method ILF + OPT-RM and other additional methods (see \S \ref{sec:ILF_OPT} for a description of ILF + OPT-RM and \S \ref{sec:main_results} for a discussion of the results). We conduct the same evaluation as described in \S \ref{sec:evaluation_results}, i.e. in a human evaluation, annotators rank various test summaries based on quality. We then calculate the win rate against human written summaries, which we use as an evaluation metric. Importantly, all methods evaluated here are trained on datasets with 5K samples. Note that the methods compared here are not exactly the same as the methods compared in Fig.~\ref{fig:finetuned_methods_comparison}. Concretely, the test summaries generated by the methods finetuning on refinements (ILF), finetuning on human summaries, and OPT-RM best-of-64 FeedME are the same as in Fig.~\ref{fig:finetuned_methods_comparison}, for the test summaries generated by corresponding methods trained on 5K samples. Here, however, we don't evaluate FeedME and finetuning on initial summaries. However, we evaluate ILF + OPT-RM (best-of-64), our best-performing model, which we also added to Fig.~\ref{fig:finetuned_methods_comparison} for reference. We also evaluate a new method called \textit{Finetuned on Feedback + Refinements}, which we describe below.

For finetuning on feedback + refinements, we us a title, post, and summary as input and the model is trained to predict the corresponding feedback and refinement. Our motivation for this approach is that generating feedback first may improve the quality of the resulting refinements, similar to the findings of previous work on self-prompting methods \citet{saunders2022self,bai2022constitutional} and the Chain of Thought (CoT) prompting technique \citet{wei2022chain}. CoT has been shown to improve the performance of models across various tasks \citet{wei2022chain} when allowing the model to reason before answering a question. 
For finetuning on feedback and refinements, we utilize the initial summaries that were used to gather human feedback, as well as the refinements generated by our method. We use the loss  $\log p(x_1, f | \text{prompt}) + \lambda \log p(\text{prompt})$, i.e. we learn to predict the refinement and the feedback. We employ the same hyperparameters as in the finetuning on refinements algorithm (including the prompt loss weight). During testing, we require initial summaries, from which we generate feedback and refinements. As initial summaries, we use the test samples generated by FeedME (as evaluated in Figure~\ref{fig:finetuned_methods_comparison}). To ensure compatibility with the 48-token length restriction of the test summaries, we append the special end token \textit{/\ n \#\#\#} to the end of the feedback and refinements during training. At test time, we set the maximum number of tokens to generate 300, and terminate generation when the stop-word \textit{/\ n \#\#\#} appears. We then apply the same postprocessing procedure outlined in Appendix~\ref{app:postprocessing} to shorten the refinements to 48 tokens. We refer to Appendix~\ref{app:finetuning_prompts} for the exact prompt templates we used.

\begin{figure}[hb!]
    \centering    \centering \includegraphics[scale=0.5]{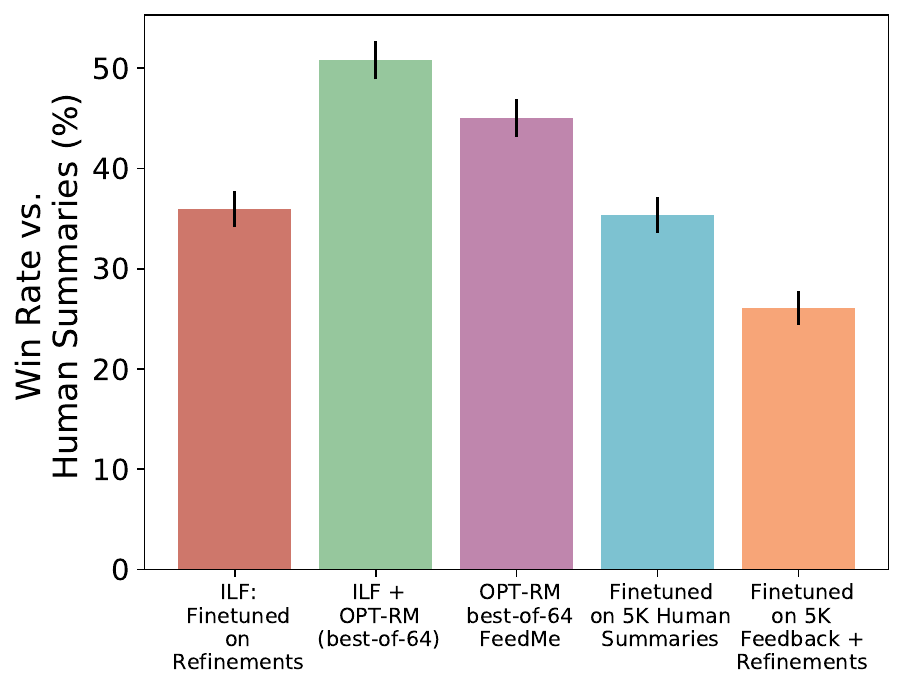}
    \caption{How often human evaluators prefer summaries from ILF: Finetuned on Refinements, OPT-RM best-of-64 FeedME, ILF + OPT-RM (best-of-64), finetuning on human summaries, and finetuning on feedback + refinements (all methods finetuned on 5K samples). ILF + OPT-RM (best-of-64) generates summaries of a similar quality to human summaries. Finetuning on feedback + refinements performs worse than finetuning on refinements (ILF).}
    \label{fig:additional_results}
\end{figure}

We present all the results in Fig.~\ref{fig:additional_results}. We find that finetuning on a set of 5K refinements achieves a win rate of $36.0 \pm 1.8 \%$, while ILF + OPT-RM (best-of-64) has a win rate of $50.8 \pm 1.9 \%$, achieving human-level summarization performance (see \S \ref{sec:main_results} for a more detailed discussion). OPT-rM best-of-64 FeedMe achieves a win rate of $45.1 \pm 1.9 \%$, finetuning on a set of 5K human-generated summaries achieves a win rate of $35.4 \pm 1.8 \%$, and finetuning on a combination of 5K feedback and refinements has a win rate of $26.1 \pm 1.7 \%$. It is worth noting that the performance of finetuning on feedback and refinements is lower than that of finetuning on refinements alone. We attribute this to the increased difficulty of generating both feedback and refinements and believe that this discrepancy may be due to limitations in our models, dataset size, or hyperparameters. Previous work has demonstrated the feasibility of training models to generate feedback \citet{saunders2022self,bai2022constitutional}, so we believe that further optimization and experimentation may improve the performance of this method. We further want to note that the results for finetuning on 5K refinements, 5K human summaries, and best-of-64 FeedME deviate from the results in Fig~\ref{fig:finetuned_methods_comparison}. This is because we compare different methods with each other, and human annotations generally contain some amount of noise (given that different people annotate the same samples).

\subsection{Multiple Iterations of ILF}
\label{app:ilf_multiple_iteratins}
Our experiments suggest that ILF is an effective method for leveraging language feedback in the training of LMs. Here we explore ILF in its most general form by doing multiple iterations of refining-and-finetuning.

\paragraph{Dataset Improvement.}
In this experiment, we evaluate the effectiveness of iterative refinement of the dataset distribution using ILF. To this end, we first finetune GPT-3-175B on 100 refinements from iteration 1 of ILF (i.e. doing one iteration of refining initial summaries, as we did in the main results of our paper, see \S \ref{sec:evaluation_results}) and refer to this finetuned model as $M_1^{100}$. The notation we use here is that the subscript indicates the iteration of ILF that the refinements were generated in, and the superscript indicates the number of overall samples the model is finetuned on. We also refer to the dataset of 100 refinements from iteration 1 as $\mathcal{D}_1^{100}$. As a baseline, we finetune $M_1^{100}$ on an additional 100 refinements from ILF iteration 1, resulting in $M_1^{200}$, i.e., a model trained on 200 refinements from ILF iteration 1. We then compare this baseline to two iterations of ILF. Specifically, we use $M_1^{100}$ to generate summaries for an additional 100 samples (the same Reddit posts as for the baseline) and collect human feedback on those summaries. We then use this feedback to generate 5 refinements using the FeedME\footnote{Ideally, one would use the same model $M_1^{100}$ to generate the refinements. However, in our case, this is not possible since we finetuned GPT-3-175B, which is not an instruction-finetuned model.} and then select the best refinement using our InstructRM method. We refer to these 100 selected refinements from the second iteration of ILF as $\mathcal{D}_2^{100}$. Finally, we finetune $M_1^{100}$ on $\mathcal{D}_2^{100}$ to obtain the model $M_{1,2}^{200}$, which has been trained on a total of 200 refinements generated in both the first and second iterations of ILF. All finetuning was performed using the same hyperparameters as described in Appendix~\ref{app:hp_tuning} for finetuning on 100 refinements. We refer to Table~\ref{tab:ilf_iterative} for an overview of all models and train datasets.

\begin{table}
\centering
\begin{tabular}{ |p{2.3cm}|p{2.3cm}||p{2.3cm}|p{2.3cm}|p{2.3cm}|| p{2.5cm}|}
 \hline
  \multirow{2}{*}{Initial Model} & \multirow{2}{*}{Finetuned Model} & \multicolumn{3}{c||}{Finetuning dataset} & \multirow{2}{*}{Produces Dataset}\\
  \cline{3-5}
  & &  ILF iteration 1 &  ILF iteration 2 &  ILF iteration 3 & \\
 \hline
 \hline
 %GPT-3-175B &  &      & & & $\mathcal{D}^{100}_1$, $\mathcal{D}^{200}_1$, $\mathcal{D}^{300}_1$ \\
%\hline
 GPT-3-175B & $M^{100}_1$   &  $\mathcal{D}^{100}_1$    & & & $\mathcal{D}^{100}_2$  \\
 \hline
  $M^{100}_1$ & $M^{200}_1$   &  $\mathcal{D}^{100*}_1$,      & & &  \\
 \hline
  GPT-3-175B & $M^{200}_{scratch,1}$  &  $\mathcal{D}^{200}_1$ & & &  \\
 \hline
  GPT-3-175B & $M^{300}_{scratch,1}$  &  $\mathcal{D}^{300}_1$ & & &  \\
 \hline
  $M^{100}_1$ & $M^{200}_{1,2}$   &  $\mathcal{D}^{100}_1$    & $\mathcal{D}^{100}_2$ &  & $\mathcal{D}^{100}_3$  \\
 \hline
  $M^{200}_{1,2}$  & $M^{300}_{1,2,3}$   &  $\mathcal{D}^{100}_1$    & $\mathcal{D}^{100}_2$ & $\mathcal{D}^{100}_3$ & \\
 \hline
    GPT-3-175B & $M^{200}_{scratch,1,2}$   &  \multicolumn{2}{c|}{$\mathcal{D}^{100}_1$ +  $\mathcal{D}^{100}_2$} & & \\
 \hline
      GPT-3-175B & $M^{300}_{scratch,1,2,3}$   &  \multicolumn{3}{c||}{$\mathcal{D}^{100}_1$ +  $\mathcal{D}^{100}_2$ + $\mathcal{D}^{100}_3$} &  \\
 \hline
\end{tabular}
\caption{Datasets (refinements) over which the models $M$ are trained, and which they generate. The superscript indicates the number of samples, whereas the subscript indicates the ILF step. In this figure we do not show FeedME which is used to generate the refinements given feedback. \newline
* these samples are new samples from the interval [100,200] of  $\mathcal{D}^{200}_1$}.
\label{tab:ilf_iterative}
\end{table}

\begin{figure*}[t!]
\begin{minipage}[t]{.45\textwidth}
 \begin{center} 
    \small{
    \cblock{150}{199}{157} ILF ($M_{1,2}^{200}$)\quad \\}
    \end{center}
    \centering
\includegraphics[scale=0.5]{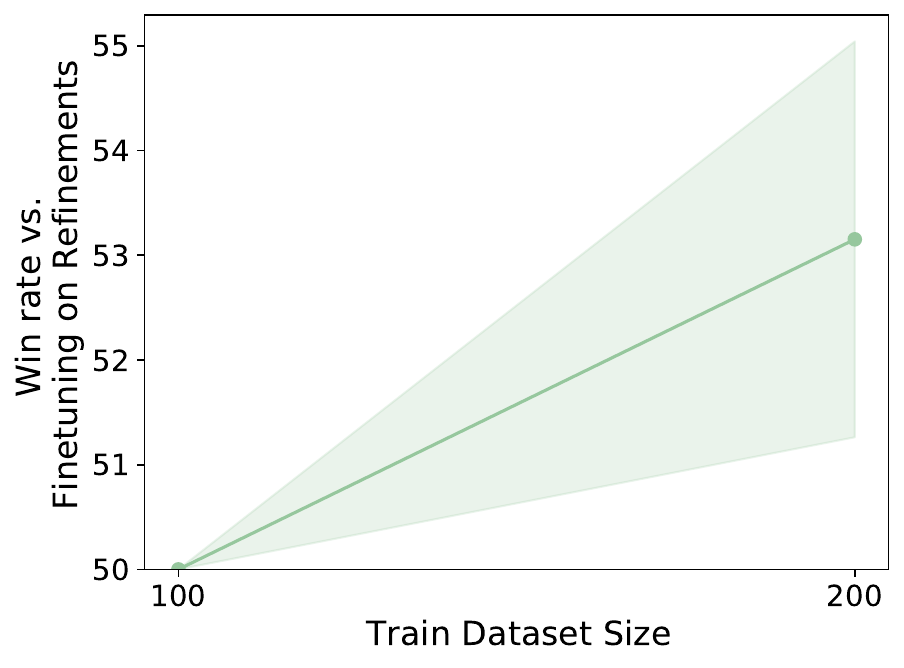}
\end{minipage}
\hfill
\begin{minipage}[t]{.45\textwidth}
\begin{center} 
    \small{
    \cblock{31}{119}{180} ILF (continuous finetuning) - 
  ($M_{1,2}^{200}$/$M_{1,2,3}^{300}$)\quad \\
     \cblock{255}{127}{14} Approximating ILF  - ($M_{scratch 1,2}^{200}$ /$M_{scratch 1,2,3}^{300}$)\quad \\
    }
    \end{center}
     \centering
\includegraphics[scale=0.5]{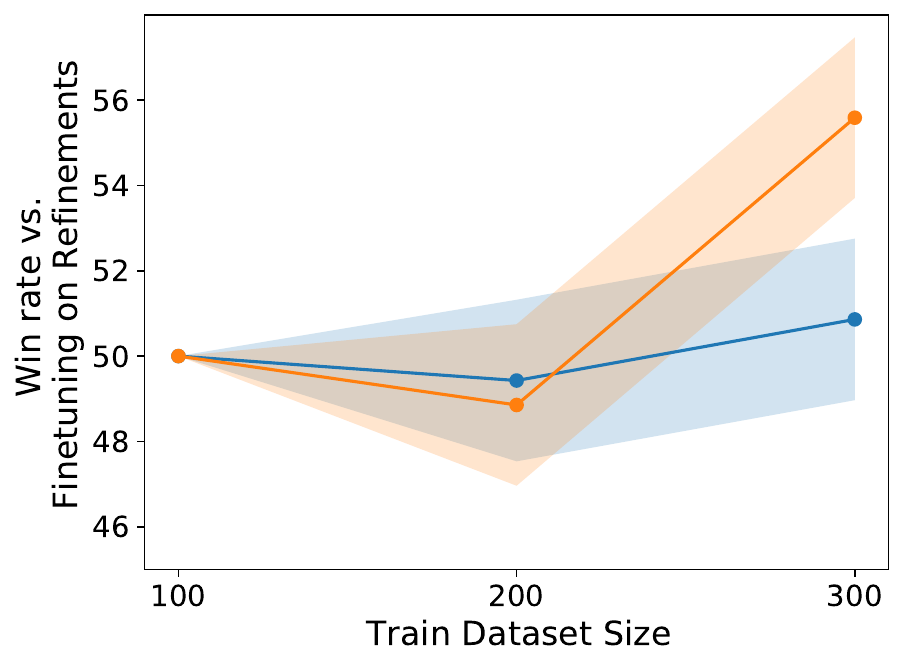}
\end{minipage}
\caption{\textbf{Left}: Win rate of 2 iterations of ILF against finetuning on the same number of refinements from the first iteration of ILF. \textbf{Right}: Win rate of 3 iterations of ILF, and approximating 3 iterations of ILF by finetuning from scratch, against finetuning on the same number of refinements from the first iteration of ILF.}
\label{fig:exert_iteration}
\end{figure*}

In this human evaluation, we compare the performance of the summaries generated by the baseline model ($M_1^{200}$) with those generated by two iterations of ILF ($M_{1,2}^{200}$) on our test set. Human evaluators are asked to indicate their preferred summary for each comparison, and the win rate of $M_{1,2}^{200}$ against $M_1^{200}$ is calculated and plotted in Fig.~\ref{fig:exert_iteration} (left)\footnote{Note, we set the win rate manually to  $50 \%$ at 100 samples, since the baseline is equivalent to one iteration of ILF.}. Our results show that two iterations of ILF outperform one iteration with a win rate of $53.2 \pm 1.9 \%$ indicating that applying multiple rounds of ILF can improve the data distribution. However, we also want to investigate whether multiple rounds of ILF lead to better models than directly finetuning on the same number of refinements from the first round from scratch. In other words, while our current baseline consists of further finetuning $M_1^{100}$ on an additional 100 samples, it is also possible to directly finetune GPT-3-175B on 200 refinements from the first iteration of ILF from scratch, i.e. $M_{scratch,1}^{200}$. We aim to determine the relative effectiveness of these two approaches in improving model performance on the text summarization task.

\paragraph{Model Improvement.}
In this experiment, we aim to compare the performance of multiple rounds of ILF to directly finetuning on a comparable number of refinements from the first iteration of ILF. As a baseline, we finetune GPT-3-175B on 200 and 300 refinements from the first iteration of ILF and conduct hyperparameter tuning as described in the Appendix~\ref{app:hp_tuning}. We then compare these baselines to two and three rounds of ILF. For the two-round ILF model, we use the previously described $M_{1,2}^{200}$. To obtain the three-round ILF model, we use $M_{1,2}^{200}$ to generate summaries for an additional 100 samples (on the same Reddit posts as for the baseline), gather human feedback, generate 5 refinements with GPT-3-175B using the feedback, and select the best refinement using InstructRM, resulting in $\mathcal{D}_3^{100}$. We then finetune $M_{1,2}^{200}$ on $\mathcal{D}_3^{100}$ to obtain the model $M_{1,2,3}^{300}$. It is important to note that while our baselines finetune GPT-3-175B from scratch on 200 and 300 refinements, the models $M_{1,2}^{200}$ and $M_{1,2,3}^{300}$ are obtained by continuously finetuning a model iteratively on additional refinements. This difference in approach may introduce a discrepancy in the results, as we use different hyperparameters, and the dataset size may affect the learning dynamics. To control for this potential difference, we also finetune GPT-3-175B from scratch on the refinements generated through various iterations of ILF. Specifically, as an alternative to $M_{1,2}^{200}$, we finetune GPT-3-175B from scratch on a concatenation of 100 refinements from the first round of ILF (i.e., $\mathcal{D}_1^{100}$) and 100 refinements from the second round of ILF (i.e., $\mathcal{D}_2^{100}$), and refer to the resulting model as $M_{scratch 1,2}^{200}$. Similarly, as an alternative to $M_{1,2,3}^{300}$, we finetune GPT-3-175B from scratch on a concatenation of 100 refinements from the first round of ILF ($\mathcal{D}_1^{100}$), 100 refinements from the second round of ILF $\mathcal{D}_2^{100}$, and refinements from the third round of ILF (i.e. $\mathcal{D}_3^{100}$), and refer to the resulting model as $M_{scratch 1,2,3}^{300}$. It is worth noting that the refinements from the second and third rounds of ILF (i.e. $\mathcal{D}_2^{100}$ and $\mathcal{D}_3^{100}$) are based on summaries generated using models that were continuously finetuned (i.e. $M_1^{100}$ and $M_{1,2}^{200}$). As such, the models $M_{scratch 1,2}^{200}$ and $M_{scratch 1,2,3}^{300}$ are not a direct application of ILF, but rather an approximation of the distribution induced by ILF. We refer to Table ~\ref{tab:ilf_iterative} for an overview of all models and train datasets.

Using a human evaluation, we compare the performance of the three methods on the test dataset: the baseline, ILF with continuous finetuning, and ILF approximated by finetuning from scratch. The results are shown in Fig.~\ref{fig:exert_iteration} (right). With this more realistic baseline, we find that directly applying ILF does not improve upon the baselines, with win rates of $49.4 \pm 1.9 \%$ and $50.9 \pm 1.9 \%$ for 200 and 300 samples, respectively. However, approximating ILF by finetuning from scratch on the distributions induced by ILF significantly improves upon the baseline for 300 samples, with a win rate of $55.6 \pm 1.9 \%$. The method is slightly worse than the baseline for 200 samples, with a win rate of $48.9 \pm 1.9 \%$. We currently hypothesize that continuous finetuning may lead to catastrophic forgetting, while finetuning from scratch may not have this problem. This could explain why $M_{scratch 1,2,3}^{300}$ performs significantly better than $M_{1,2,3}^{300}$ for 300 samples. Specifically, $M_{1,2}^{200}$ may actually generate an improved distribution in the third iteration of ILF. However, when further finetuning $M_{1,2}^{200}$ on this improved distribution $\mathcal{D}_2^{100}$, the model may forget what it learned previously. On the other hand, the model $M_{scratch 1,2,3}^{300}$ that learns from scratch on the concatenation of all datasets produced by ILF may actually benefit from the improved dataset distribution because it does not unlearn anything. It is, however, unclear why $M_{scratch 1,2}^{200}$ does not benefit from the improved data distribution $\mathcal{D}_2^{100}$. It is also possible that the hyperparameters play a significant role in the final performance of the various models and that the dataset size has a strong influence on model performance (e.g., finetuning on more samples may be more stable than finetuning on fewer samples). In future work, we plan to conduct more elaborate experiments to answer these questions and better understand the effects of the dataset size and number of iterations on ILF. Specifically, we aim to run multiple iterations of ILF and use $M_{scratch 1,2}^{200}$ as the model to generate summaries in the third round of ILF (instead of $M_{1,2}^{200}$). This would be a direct implementation of ILF, rather than an approximation of it, as we would be finetuning the same model with which we are also generating an improved distribution. We also hope to investigate the effect of the dataset size and number of iterations on ILF. Overall, our results suggest that ILF has the potential to improve the performance of natural language processing systems by continuously incorporating human feedback into the training of language models, but further research is needed to fully understand the best ways to leverage this approach.

\subsection{Part-of-Speech Distribution for Finetuning Datasets}
\label{app:pos}
\begin{figure*}[t!]
\begin{minipage}[t]{.45\textwidth}
    \centering
\includegraphics[scale=0.4]{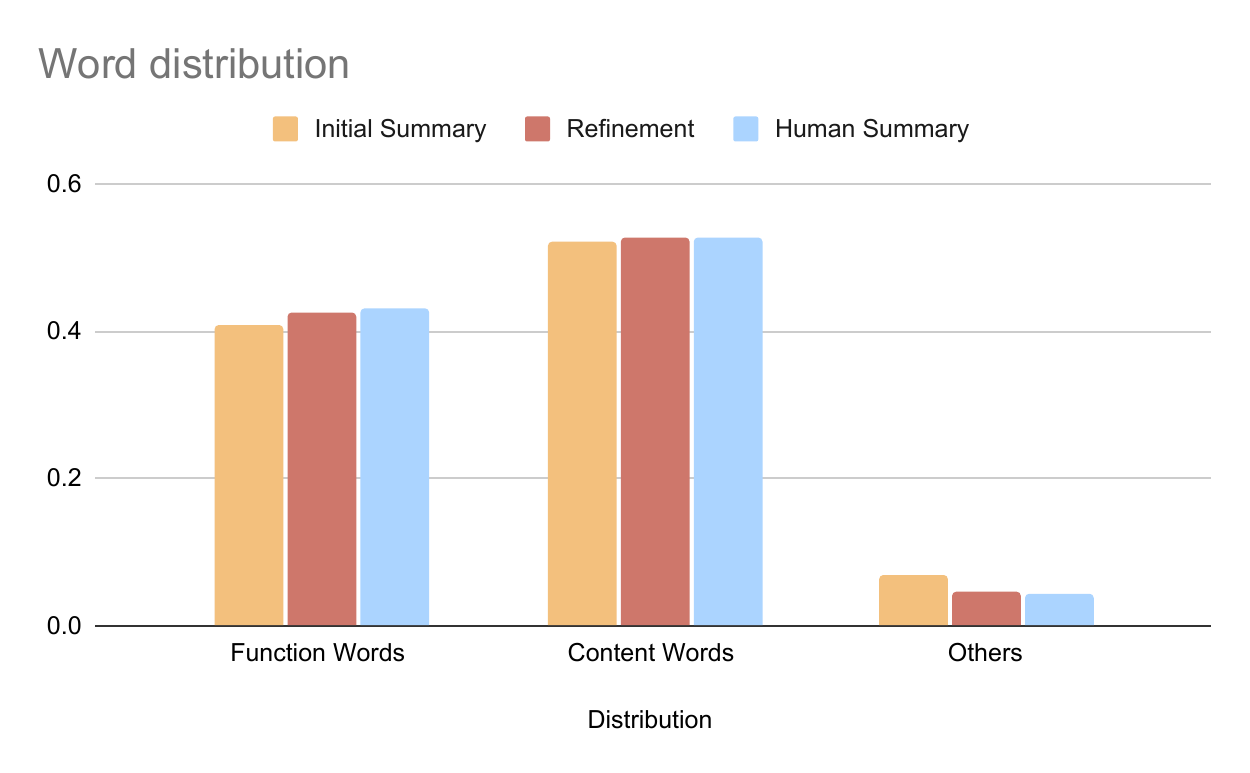}
\end{minipage}
\hfill
\begin{minipage}[t]{.45\textwidth}
     \centering
\includegraphics[scale=0.4]{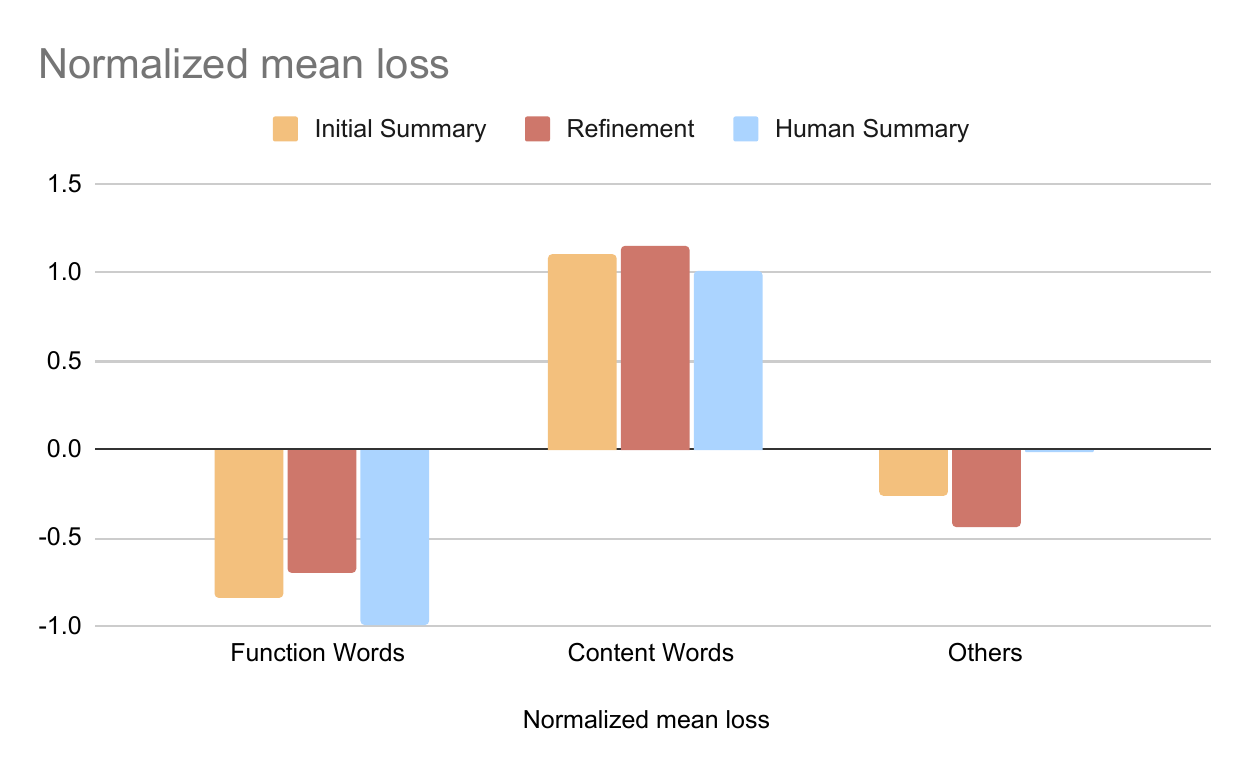}
\end{minipage}
\caption{Distribution of tokens of various finetuning datasets with 1K samples in terms of content and function words. We only evaluate the various completions, i.e., summaries, since the prompts are the same for all distributions.}
\label{fig:pos_distribution}
\end{figure*}

We evaluate the negative log-likelihood of GPT-3-175B on the three finetuning datasets, i.e. on initial summaries, refinements, and human summaries. We use the training dataset with 1K samples and calculate the negative log-likelihood over different Part-of-Speech tags. We use Stanza \cite{qi2020stanza} as the PoS tagger for this experiment and then we separate the words into three groups: function words, content words, and others. The function words are words that have little lexical meaning: articles, pronouns, adpositions, conjunctions, auxiliary verbs, particles and interjections. On the other hand, content words are words that contain semantic information: nouns, adjectives, adverbs and lexical verbs. We keep numbers and symbols under the group \textit{others}. With this analysis, we want to spot different patterns between model-generated (initial summaries and refinements) and human-written summaries. Note that a high negative log-likelihood implies a high loss. We present the results in Fig~\ref{fig:pos_distribution}. Since the average loss is higher for human summaries, we normalize all the loss values by transforming them to have mean 0 and standard deviation 1. Overall, the word distribution is very similar for all three finetuning datasets. In terms of normalized mean loss, it is interesting how the content words have a bigger influence on the refinements dataset. We believe that this is related to our results in section \ref{sec:main_results}, where we obtain the best results when finetuning on refinements. 

\subsection{Comparison to Results of \citet{scheurer2022training} }
\label{app:comparison_scheurer_22}

Here we relate our results to previous work by \citet{scheurer2022training}. In Fig. 2 of \citet{scheurer2022training}, they compare their method of finetuning on refinements against various baselines, such as finetuning on initial summaries, sampling from FeedME (called InstructGPT), and sampling from GPT-3-175B. They calculate the win rate of all methods against human written summaries \citep{volske-etal-2017-tl} that are automatically extracted from Reddit. As shown in \S\ref{sec:data} and App.\ref{sec:dataset_analysis}, our human summaries are preferred $72.3 \pm 3.2 \%$ to the human summaries of \citet{volske-etal-2017-tl}. This implies that the win rates in \citet{scheurer2022training} are much higher than in our case since we use a much stronger baseline. 

We now present three differences between the results found in \citet{scheurer2022training} and the results found in our paper. Then we will provide various potential reasons that could explain the differences. First, when comparing the results (in relative terms) in \citet{scheurer2022training} Fig. 2 to our results in Fig.~\ref{fig:finetuned_methods_comparison} where we finetune on 100 samples, we see differences in performance. \citet{scheurer2022training} reports that finetuning on refinements outperforms finetuning on initial summaries. And both methods outperform sampling from FeedME (i.e., InstructGPT). In our experiments finetuning on 100 refinements achieves a win rate of $19.6 \pm 1.5 \%$ against human summaries, finetuning on initial summaries a win rate of $19.6 \pm 1.5 \%$, and FeedME a win rate of $20.8 \pm 1.5 \%$. Thus both finetuned methods perform equally and are worse than sampling from FeedME. 

Second, we compare the results of refining a summary with feedback. Note that \citet{scheurer2022training} uses an embedding-based scoring function to select refinements, whereas we use InstructRM. In \citet{scheurer2022training} Fig. 3 (left) \textsc{Refine with Feedback + Best of N} achieves a win rate of $67.0 \pm 3.1 \%$ against initial summaries (sampled from FeedME), \textsc{Refine with Feedback} achieves a win rate of $60.5 \pm 3.0 \%$, \textsc{Refine without Feedback} achieves $50.3 \pm 2.6 \%$ and Human Summaries have a win rate of $60.8 \pm 3.4$. In our Fig.~\ref{fig:win_rates_and_incorporating_feedback} (left) Refine with Feedback + Best-of-5 achieves a win rate of $69.1 \pm 1.9 \%$, Refine with Feedback achieves a win rate of $63.9 \pm 2.0 \%$, Refinement without Feedback achieves a win rate of $59.4 \pm 2.0 \%$ and Human Summaries a win rate of $83.2 \pm 1.7 \%$. The difference in the human summaries is expected, given that we use better human summaries. The Refinement without Feedback method achieves higher results in our work than in \citet{scheurer2022training}. 

Third, it is also noteworthy that using the embedding similarity as a scoring function worked well in \citet{scheurer2022training}, while it does not work in our setting (see Table \ref{tab:scoring_function_results} and \S\ref{sec:scoring_function_results} for a discussion of the results). We believe this is because the feedback we collect is written by many annotators and is thus much more diverse, while in \citet{scheurer2022training}, the authors themselves
wrote the feedback.

Here we now list various differences in the setup of \citet{scheurer2022training} and our paper, which could all account for the different results.
\begin{enumerate}
    \item \citet{scheurer2022training} use an embedding similarity as a scoring function, while we use InstructRM Ensemble. Looking at Tab.~\ref{tab:scoring_function_results} and the corresponding discussion in  \S\ref{sec:scoring_function_results}, already shows that the methods are very different. 
    \item The human-written summaries are of much higher quality in our paper than in \citet{scheurer2022training} (see \S\ref{sec:data} and App.~\ref{sec:dataset_analysis})

    \item In \citet{scheurer2022training}, the annotation instructions specifically state that the feedback should mention how to improve a summary. In our work, we collect much more unrestricted and diverse feedback. This difference is also apparent in the fact that the embedding similarity does not work well as a scoring function in our setting. 
    \item In \citet{scheurer2022training}, the authors themselves annotated the data, i.e., they wrote the feedback and evaluated the final summaries. In our case, we use independent evaluators who are trained on this task. Using 31 annotators overall also gives us a more diverse and less biased estimate of our methods. Also, doing human evaluations is inherently noisy and will never lead to the exact same results. 
    \item The evaluation in \citet{scheurer2022training} was done on a different dataset than in this work. Specifically, they used only 100 samples to evaluate their method, while we use a test set of 698 samples.
    \item The hyperparameters in \citet{scheurer2022training} used for sampling and finetuning are different from the hyperparameters used in our work.
    \item Overall, we use different prompts than \citet{scheurer2022training} (see App.~\ref{app:finetuning_prompts} and App.~\ref{app:summarization_prompts})
\end{enumerate}

\section{Annotator Instructions}
\label{app:annotator_instructions}

Overall we completed many annotations to create datasets and evaluate our algorithm. The instructions were task-specific and also continuously updated. In the following, we provide the instructions we used to create our train dataset and the instructions we provided for evaluating the summary quality (of 6 summaries). We will not share more instructions for brevity but can provide them upon request.

\subsection{Train Dataset Annotation Instructions}
\label{app:train_data_annotation_instructions}

\textbf{Task Overview}

You are given a Reddit Post, which you first need to read carefully. You then need to complete 5 subtasks which consist of comparing two summaries, writing feedback on a summary, classifying the type of feedback, indicating whether there is additional Feedback, and writing an ideal summary. When doing these tasks, please adhere to the guidelines below.

\textbf{What makes for a good summary?} Roughly speaking, a good summary is a short piece of text that has the essence of the original text. A good summary tries to accomplish the same purpose and conveys the same information as the original text. We would like you to consider these different dimensions of summaries:

\textbf{Essence}: Is the summary a good representation of the post? How well does the summary cover the important information in the post?

\textbf{Clarity}: Is the summary reader-friendly? Does it express ideas clearly? 

\textbf{Accuracy}: Does the summary contain the same information as the post? 

\textbf{Purpose}: Does the summary serve the same purpose as the original post? 

\textbf{Concise}: Is the summary short and to the point?

\textbf{Style}: Is the summary written in the same style as the original post?

Generally speaking, we give higher weight to the dimensions at the top of the list. The evaluation can be complicated though, since none of the above dimensions are simple yes/no matters, and there aren’t hard and fast rules for trading off different dimensions. Use your best judgment and common sense to make these trade-offs. In case the subreddit, title, and Reddit post leave open some ambiguity about what happened, it is important to accurately reflect that in your annotations and not just interpret the text in a certain way. Always look at all the subreddit, title, and Reddit Post and use all information given to make your judgments (sometimes the title may contain crucial information that does not appear in the post but should nevertheless be used).

First, read the Subreddit category, title, and post carefully. A Subreddit is a forum dedicated to a specific topic on the website Reddit. Take your time with this step and re-read the parts that you might not have understood at first. Below is a detailed description of the task you will need to complete for each Reddit post.

Below is a detailed description of each task you will need to complete for each Reddit post: 

\begin{enumerate}
    \item \textbf{Comparison Task}: Given a pair of summaries, indicate which is better.
    
    \textit{Details}: Use the above description of what makes a good summary. It is alright to choose either summary if both summaries are identical copies of each other or if there is no distinguishing feature that makes one summary superior to the other. However, if there is a small detail that makes one summary better than the other, that is enough reason to select that summary.

    \item \textbf{Feedback Task}: Write short and simple feedback on the given summary about the single, most important shortcoming of the summary. The feedback should NOT mention what category (Accuracy, Coverage, Coherence, other) the feedback belongs to, nor should it assume knowledge about the definitions of “Coverage”, “Accuracy”, or “Coherence” (see below). Otherwise, the feedback should be as short and simple as possible while still addressing the most important shortcoming of the summary.

    \textit{Details}: You can write the feedback in one or several sentences, but it should only address the single, most important shortcoming of the summary and be as short as possible. There are no other restrictions as to how you write the feedback and what exactly it addresses. If there are no shortcomings in the summary, the feedback can also mention a positive thing about the summary.
    Use the description of what makes a good summary to trade off the various dimensions that make for a good summary. Often the feedback will (but does not have to) address one of the following axes.  
    \begin{itemize}
        \item \textbf{Coverage}: For this axis, answer the question, “how well does the summary cover the important information in the post?” A summary has good coverage if it mentions the main information from the post that’s important to understand the situation described in the post. A summary has poor coverage if someone reading only the summary would miss several important pieces of information about the situation in the post. A summary with good coverage should also match the purpose of the original post (e.g., to ask for advice).
        \item \textbf{Accuracy}: For this axis, answer the question, “does the factual information in the summary accurately match the post?” A summary is accurate if it doesn’t say things that aren’t in the article, doesn’t mix up people, and is generally not misleading.  If the summary says anything at all that is not mentioned in the post or contradicts something in the post, it is NOT accurate.
        \item \textbf{Coherence}: For this axis, answer the question, “how coherent is the summary on its own?” A summary is coherent if, when read by itself, it’s easy to understand and free of English errors. A summary is not coherent if it’s difficult to understand what the summary is trying to say. Generally, it’s more important that the summary is understandable than being free of grammar errors.
    \end{itemize}
    
    Additional Rules: 
    The feedback should NOT mention what category (Accuracy, Coverage, Coherence, other) the feedback belongs to, nor should it assume knowledge about the definitions of “Coverage”, “Accuracy”, “Coherence”, or “other” (as defined above).
    Example: One should NOT write "This is missing in the area of coverage", or "This summary lacks in the category of accuracy, because ...". The feedback should be understandable to a person who has never read the definition of "Coverage", "Accuracy", and "Coherence". You are, however, ALLOWED to use those words if they make sense on their own, e.g., you CAN say, "This summary does not cover the important parts of the text because", or "This summary is inaccurate as it states ...", or "This is not a coherent summary because ...".

    \item \textbf{Feedback Type Task}: If your feedback falls into the categories Accuracy-related, Coherence-related, or Coverage-related, mark it as such by checking the corresponding checkbox for the (single) category it is related to. If your feedback is not related to any of these three categories, then check the "Other" checkbox.

    \item \textbf{More Feedback Task}: Answer with Yes if there is additional Feedback about an important shortcoming of the summary that you would want to mention and No otherwise.

    \item \textbf{Ideal Summary Task}: Ideal Summary Task: Write a short summary for the Reddit post that is ideal in your view.

    \textit{Details}: The ideal summary should be ideal in terms of all the criteria mentioned above, i.e., essence, clarity, accuracy, coverage, purpose, conciseness, coherence, and style. In other words, you should not be able to find an obvious critique of the ideal summary that you write. It is okay to reuse parts of previous summaries but only if those parts should be a part of an ideal summary. The ideal summary should maximally be 48 tokens long (otherwise, you can't submit your annotation). Tokens are generated by taking your ideal summary and splitting up certain words into individual pieces (this is necessary to train our AI). The interface will show you how many tokens your ideal summary has already taken up.
\end{enumerate}

\subsection{Summary Quality Evaluation Instructions}
\label{app:summary_quality_instructions}

\textbf{Task Overview}

You will be given a Subreddit category, a title, and a Reddit Post, which you first need to read carefully. Your task is then to compare 6 summaries and rank them according to quality.

\textbf{What makes for a good summary?} Roughly speaking, a good summary is a short piece of text that has the essence of the original text. A good summary tries to accomplish the same purpose and conveys the same information as the original text. We would like you to consider these different dimensions of summaries:

\textbf{Essence}: Is the summary a good representation of the post? How well does the summary cover the important information in the post?

\textbf{Clarity}: Is the summary reader-friendly? Does it express ideas clearly? 

\textbf{Accuracy}: Does the summary contain the same information as the post? 

\textbf{Purpose}: Does the summary serve the same purpose as the original post? 

\textbf{Concise}: Is the summary short and to the point?

\textbf{Style}: Is the summary written in the same style as the original post?

Generally speaking, we give higher weight to the dimensions at the top of the list. The evaluation can be complicated though, since none of the above dimensions are simple yes/no matters, and there aren’t hard and fast rules for trading off different dimensions. Use your best judgment and common sense to make these trade-offs. In case the subreddit, title, and Reddit post leave open some ambiguity about what happened, it is important to accurately reflect that in your annotations and not just interpret the text in a certain way. Always look at all the subreddit, title, and Reddit Post and use all information given to make your judgments (sometimes the title may contain crucial information that does not appear in the post but should nevertheless be used).

First, read the Subreddit category, title, and post carefully. A Subreddit is a forum dedicated to a specific topic on the website Reddit. Take your time with this step and re-read the parts that you might not have understood at first. Below is a detailed description of the task you will need to complete for each Reddit post.

\textbf{Comparison Task}: Given 6 summaries, indicate which is better by ranking them according to quality. Rank 1 is considered the highest rank, and Rank 6 is considered the lowest rank. The summary with the best quality should be ranked highest, i.e., as Rank 1, and the summary with the worst quality should be ranked lowest, i.e. Rank 6. Use the above description of what makes a good summary. Ties between summaries are allowed, but only if summaries are exact copies of each other or if there is no distinguishing feature that makes one summary superior to the other. However, if there is a small detail that makes one summary better than the other, that is enough reason to rank that summary as better than the other summary. We use Standard Competition ranking (i.e., example rankings of 122456). In standard competition ranking, items that compare equally receive the same ranking number, and then a gap is left in the ranking numbers. The number of ranking numbers that are left out in this gap is one less than the number of items that are compared equally. Equivalently, each item’s ranking number is 1 plus the number of items ranked above it.

\section{Prompts}
\subsection{Summarization Prompts}
\label{app:summarization_prompts}
We report all prompt templates used to generate \textsc{Intial Summaries}, \textsc{Refinement with Feedback}, and \textsc{Refinement without Feedback} in Table~\ref{tab:summarization_prompt_template}.

\begin{table*}[ht]
    \centering
    \begin{tabular}{p{4cm} p{8cm} c}
    \toprule
    \textbf{Methods} &  \textbf{Format} \\ 
  \hline
    \textsc{Initial Summary} & Write an excellent summary of the given text.  &  \\
    & \\
    & Title: \{\texttt{title}\}  \\
    & \\
    & Text: \{\texttt{text}\} & \\
    & \\
    & TL;DR:  & \\ 
   \hline
 \textsc{Refinement with Feedback} & Write an excellent summary that incorporates the feedback on the given summary and is better than the given summary.  &  \\
        & \\
        & Title: \{\texttt{title}\}  \\
        & \\
        & Text: \{\texttt{text}\} & \\
            & \\
        & Summary: \{\texttt{summary}\} & \\
            & \\
        & Feedback on Summary: \{\texttt{feedback}\} & \\
            & \\
        & Improved TL;DR:  & \\ \hline
  \textsc{Refinement without Feedback} & Write an excellent summary that is better than the given summary.  &  \\
      & \\
        & Title: \{\texttt{title}\}  \\
            & \\
        & Text: \{\texttt{text}\} & \\
            & \\
        & Summary: \{\texttt{summary}\} & \\
            & \\
        & Improved TL;DR:   & \\
\bottomrule
    \end{tabular}
    \caption{Prompt templates used for summarization.}
    \label{tab:summarization_prompt_template}
\end{table*}

\subsection{InstructRM Prompts}
\label{app:instruct_rm_prompts}
We instructed one of the authors of this paper (who at the time had not been involved in the research project) to write 5 prompts that would achieve the goal of selecting high-quality summaries, i.e., refinements. The author did not have any domain knowledge or prior information on what kinds of prompts would work. The instructions provided to the author can be viewed \href{https://docs.google.com/document/d/1J1wb7JJLDHS1eu2n20t5CQtw7HBKy7jf6N2tN9nUJKU/edit?usp=sharing}{here}. We report all 5 prompt templates in Table~\ref{tab:instruct_rm_prompts}.

\begin{longtable}{p{4cm} p{8cm} c}
\toprule
    \textbf{InstructRM Prompts} &  \textbf{Format} \\ 
  \hline
    \textsc{Prompt 1} & Here's a summary of a Reddit post, feedback on the summary, and a new summary. You will be asked to determine whether the new summary incorporates the feedback provided. \\
    & \\
    & A good summary is a short piece of text that has the essence of the original text. A good summary tries to accomplish the same purpose and conveys the same information as the original text.  &  \\
    & \\
    & Post title: \{\texttt{title}\}  \\
    & \\
    & Below, there's the content of the post that was summarized. & \\
    & \\
    & Original post: \{\texttt{text}\} & \\
    & \\
    & Original summary: \{\texttt{summary}\} & \\
    & \\
    & A human then provided feedback on the above summary.& \\
    & \\
    & Feedback: \{\texttt{feedback}\} & \\
    & \\
    & Based on this feedback, a new summary was written. & \\
    & \\
    & New summary: \{\texttt{refinement}\} & \\
    & \\
    & Does this new summary incorporate the feedback provided? Answer Yes or No. \\
    & \\
    & Answer:  \\ 
   \hline
   
    \textsc{Prompt 2} & Post title:  \{\texttt{title}\} &\\
        & \\
    & Original post: \{\texttt{text}\} & \\
    & \\
   & Original summary:  \{\texttt{summary}\} \\
   & \\
   & Feedback:  \{\texttt{feedback}\} \\
   & \\
      & New summary:  \{\texttt{refinement}\} \\
   & \\
   & Question: Does the new summary incorporate the feedback provided? Answer Yes or No.\\
   & \\
   & Answer: \\
   \hline

    \textsc{Prompt 3}  & You will be given a Reddit post title, its content, an original summary of that post, and feedback for that summary. Then, your goal will be to determine whether the new summary improves upon the original with respect to provided feedback. & \\
    & \\
    & Post title:  \{\texttt{title}\} &\\
    & \\
    & Post content: \{\texttt{text}\} & \\
    & \\
   & Original summary:  \{\texttt{summary}\} \\
   & \\
   & Feedback:  \{\texttt{feedback}\} \\
   & \\
      & New summary:  \{\texttt{refinement}\} \\
   & \\
   & Question: Does the new summary incorporate the feedback provided? Answer True or False.\\
   & \\
   & Answer: \\
   \hline

    \textsc{Prompt 4}  & Here's a summary of a Reddit post, feedback on the summary, and a new summary. You will be asked to determine whether the new summary incorporates the feedback provided. & \\
    & \\
    & A good summary is a short piece of text that has the essence of the original text. A good summary tries to accomplish the same purpose and conveys the same information as the original text. Remember, you will be asked to determine whether the new summary incorporates the feedback provided. & \\
    & \\
    & Post title:  \{\texttt{title}\} &\\
    & \\
    & Below, there's the content of the post that was summarized.  \\
    & \\
    & Original Post: \{\texttt{text}\} & \\
    & \\
    & Remember, you will be asked to determine whether the new summary incorporates the feedback provided. Here's the original summary. \\
    & \\
   & Original summary:  \{\texttt{summary}\} \\
   & \\
   & Remember, you will be asked to determine whether the new summary incorporates the feedback provided. A human then provided feedback on the above summary. \\
    & \\
   & Feedback:  \{\texttt{feedback}\} \\
   & \\
   & Based on this feedback, a new summary was written. \\ 
   & \\
      & New summary:  \{\texttt{refinement}\} \\
   & \\
   & Does this new summary incorporate the feedback provided? Answer Yes or No.\\
   & \\
   & Answer: \\
   \hline

       \textsc{Prompt 5}  & Here's a summary of a Reddit post, feedback on the summary, and a new summary. You will be asked to determine whether the new summary incorporates the feedback provided. & \\
    & \\
    & The feedback was: & \\
    & Feedback: {feedback} & \\
    & \\
    & Here's the post that was summarized in the first place. \\
    & \\
    & Post title:  \{\texttt{title}\} &\\
    &\\
    & Original Post: \{\texttt{text}\} & \\
    & \\
    & Remember, you will be asked to determine whether the new summary incorporates the feedback provided. Here's the original summary. \\
    & \\
   & Original summary:  \{\texttt{summary}\} \\
   & \\
   & Remember, you will be asked to determine whether the new summary incorporates the feedback provided. A human then provided feedback on the above summary. Here's the feedback again. \\
    & \\
   & Feedback:  \{\texttt{feedback}\} \\
   & \\
   & Based on this feedback, a new summary was written. \\ 
   & \\
      & New summary:  \{\texttt{refinement}\} \\
   & \\
   & Does this new summary incorporate the feedback provided? Answer True or False.\\
   & \\
   & Answer: \\
   \hline

\bottomrule
\caption{Prompt templates used for InstructRM Ensemble.}
\label{tab:instruct_rm_prompts}
\end{longtable}

\subsection{Finetuning Prompts}
\label{app:finetuning_prompts}
In Table~\ref{tab:finetuning_prompts}, we report the prompts we use for finetuning on summaries and finetuning on feedback + refinements. The completion for finetuning on summaries indicates that we can have completions generated from various sources, i.e., either initial summaries from \textit{FeedMe}, refinements generated with our method, or ideal human written summaries. For finetuning feedback + refinements, we first generate the feedback and then the refinement.

\begin{longtable}{p{0.2\textwidth} p{0.4\textwidth} c c }
\toprule
    \textbf{Methods} &  \textbf{Prompt} & \textbf{Completion} \\ 
  \hline
    \makecell{\textsc{Finetuning on} \\ \textsc{Summaries}} & Write an excellent summary of the given text.  & \makecell{ \{\texttt{summary/refinement/human summary}\}}  
    & \\
    & Title:  \{\texttt{title}\} & \\
    & \\
    & Text:  \{\texttt{post}\} & \\
    & \\
    & TL;DR: & \\
   \hline
     \makecell{\textsc{Finetuning on} \\ \textsc{Feedback} \\ \textsc{ + Refinements}} &Write an excellent summary that incorporates the feedback on the given summary and is better than the given summary.  & \makecell{\{\texttt{feedback}\} \\ \\ Improved TL;DR: \{\texttt{refinement}\} \\ \#\#\#} & \\
    & \\
    & Title:  \{\texttt{title}\} & \\
    & \\
    & Text:  \{\texttt{post}\} & \\
    & \\
    & Summary:  \{\texttt{summary}\} & \\
    & \\
    & Feedback on summary: & \\
    & \\
   \hline
\bottomrule
\caption{Prompt templates used for Finetuning on Summaries and Feedback + Refinement.}
\label{tab:finetuning_prompts}
\end{longtable}

\subsection{Reward Model Prompts}
\label{app:rm_prompts}
\begin{longtable}{p{0.2\textwidth} p{0.4\textwidth} c c }
\toprule
    \textbf{Reward Model Type} &  \textbf{Prompt} & \textbf{Completion} \\ 
  \hline
    \makecell{\textsc{Binary RM}} & Title:  \{\texttt{title}\} & \makecell{ \{\texttt{" Yes"/" No"}\}}   & \\
    & \\
    & Text:  \{\texttt{post}\} & \\
    & \\
    & TL;DR: \{\texttt{summary\_A/summary\_B}\} & \\
    & \\
    & Question: Is the above an excellent summary of the given text? An excellent summary is coherent, accurate, concise, and detailed. Answer with Yes or No. & \\
    & \\ 
    & Answer: & \\
   \hline
       \makecell{\textsc{Comparison RM}} & Title:  \{\texttt{title}\} & \makecell{ \{\texttt{" A"/" B"}\}}   & \\
    & \\
    & Text:  \{\texttt{post}\} & \\
    & \\
    & Summary A: \{\texttt{summary\_A}\} & \\
    & \\
    & Summary B: \{\texttt{summary\_B}\} & \\
    & \\
    & Question: Which summary is the better one? An excellent summary is coherent, accurate, concise, and detailed. Answer with A or B. & \\
    & \\ 
    & Answer: & \\
   \hline
\bottomrule
\caption{Prompt templates used for training the reward model with the language model loss. Both classification and comparison prompts are shown.}
\label{tab:reward_model_prompts}
\end{longtable}

% keep this out for blind review

\end{document}